\documentclass[11pt]{article}
\usepackage[margin=1.0in]{geometry}

\usepackage{natbib}
\usepackage{graphicx}
\usepackage[table]{xcolor}
\usepackage{amsmath}    %
\usepackage{amsfonts}
\usepackage{amssymb}    %
\usepackage{amsthm}     %
\usepackage{bm}         %
\usepackage{enumitem}
\usepackage{wrapfig}
\usepackage{subcaption}
\usepackage[dvipsnames]{xcolor}
\usepackage[
   pdftex=true,
    colorlinks=true,
    linkcolor=RoyalPurple,   %
    citecolor=Blue,     %
    urlcolor=Violet     %
]{hyperref}
\usepackage[capitalise]{cleveref}
\usepackage{listings}

\newcommand{\bR}{\mathbb{R}}
\newcommand{\prox}{\operatorname{prox}}
\newcommand{\cL}{\mathcal{L}}
\newcommand{\bE}{\mathbb{E}}
\newcommand{\cN}{\mathcal{N}}
\newcommand{\KL}{{\mathrm{KL}}}

\usepackage{minted}
\setminted{
    fontsize=\small,
    breaklines=true,
    breakanywhere=true,
    breaksymbolleft=,
    breaksymbolright=,
    frame=none,
    baselinestretch=1,
} 
\def\isArxiv{1}

\title{\textsc{ProxT2I}: Efficient Reward-Guided Text-to-Image Generation \\ via Proximal Diffusion}

\author{
Zhenghan Fang\\
Johns Hopkins University\\
\texttt{zfang23@jhu.edu}
\and
Jian Zheng\thanks{This work does not relate to the author's position at Amazon.}\\
Amazon\\
\texttt{nzhengji@amazon.com}
\and
Qiaozi Gao\footnotemark[1]\\
Amazon\\
\texttt{qzgao@amazon.com}
\and
Xiaofeng Gao\footnotemark[1]\\
Amazon\\
\texttt{gxiaofen@amazon.com}
\and
Jeremias Sulam\\
Johns Hopkins University\\
\texttt{jsulam1@jhu.edu}
}
\date{}

\begin{document}
\maketitle
\vspace{-5pt}
\begin{abstract}
Diffusion models have emerged as a dominant paradigm for generative modeling across a wide range of domains, including prompt-conditional generation. The vast majority of samplers, however, rely on forward discretization of the reverse diffusion process and use score functions that are learned from data. Such forward and explicit discretizations can be slow and unstable, requiring a large number of sampling steps to produce good-quality samples.
In this work we develop a text-to-image (T2I) diffusion model based on \emph{backward} discretizations, dubbed \textsc{ProxT2I}, relying on learned and conditional proximal operators instead of score functions. We further leverage recent advances in reinforcement learning and policy optimization to optimize our samplers for task-specific rewards. Additionally, we develop a new large-scale and open-source dataset comprising 15 million high-quality human images with fine-grained captions, called \textsc{LAION-Face-T2I-15M}, for training and evaluation. Our approach consistently enhances sampling efficiency and human-preference alignment compared to score-based baselines, and achieves results on par with existing state-of-the-art and open-source text-to-image models while requiring lower compute and smaller model size, offering a lightweight yet performant solution for human text-to-image generation.
\end{abstract} %
\section{Introduction}
Generative artificial intelligence has rapidly transformed the landscape of content creation, enabling the synthesis of realistic text, images, audio, and video. In particular, diffusion models, generative models based on discretizations of stochastic differential equations (SDEs) and ordinary differential equations (ODEs), have emerged as the workhorse behind recent advances in image and text-to-image (T2I) generation \citep{ho2020denoising,song2021score}. By simulating a carefully-designed continuous-time dynamics that transforms noise into data, these diffusion-based methods achieve state-of-the-art image quality and flexible semantic controllability, underpinning many widely used systems such as Stable Diffusion \citep{rombach2022ldm}, Imagen \citep{saharia2022imagen}, and DALL·E~2 \citep{ramesh2022hierarchical}.

Despite the success of diffusion models, high-quality generation with very few sampling steps remains an open challenge. Conventional score-based samplers, based on forward, explicit discretization of continuous-time processes, often deteriorate significantly in sample quality when the number of steps is reduced, or require carefully tuned and specialized solvers to improve sampling speed \citep{salimans2022progressive,lu2022dpm,lu2025dpm}. This leads to inefficiencies in practical deployments where fast inference is essential. Efficient sampling is particularly challenging in \emph{human} image generation, where the image quality--particularly for faces and hands--remains less satisfactory, with subtle but very perceptible artifacts or unrealistic features still common \citep{liao2024facescore}. 

\begin{figure}[t]
    \centering
    \if\isArxiv 1
    \includegraphics[width=0.6\textwidth]{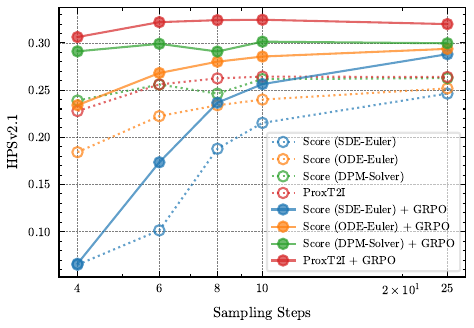}
    \fi
    \if\isArxiv 0
    \includegraphics[width=0.45\textwidth]{figures/test_curve_prompts_1024_hpsv2.1.pdf}
    \vspace{-8pt}
    \fi
    \caption{HPSv2.1 score \citep{wu2023human} vs. number of sampling steps at $256^2$ resolution. Our ProxT2I achieves more efficient and human-preference-aligned T2I generation than competing methods.}
    \if\isArxiv 0
    \vspace{-18pt}
    \fi
    \label{fig:256}
\end{figure}

An alternative approach to designing diffusion models, recently proposed by  \citep{fang2025beyond}, relies on  %
applying a \emph{backward} discretization to the reverse diffusion process. Unlike traditional forward-discretized solvers that approximate the reverse-time SDE via explicit updates based on scores, the proximal diffusion models from \citep{fang2025beyond} employ backward and implicit updates, and leverage proximal operators--instead of the gradients--of the log-density. These proximal-based solvers achieve improved theoretical convergence rates and superior empirical sampling efficiency in generating high-quality samples. So far, proximal diffusion models have been limited to unconditional generation and showcased on low-dimensional data, leaving its potential for conditional generation of high-resolution images unexplored. 

Moreover, diffusion models trained simply with denoising score-matching objectives often fall short on downstream objectives such as text-image alignment, aesthetics, safety requirements, or human preference. In this context, reinforcement learning (RL) has become a crucial post-training paradigm to optimize diffusion models for such goals \citep{black2024training,fan2023dpok,clark2023directly,uehara2024fine,uehara2024understanding}. Unfortunately, policy-gradient methods in RL typically rely on stochastic sampling paths and are therefore not directly applicable to deterministic (and faster) ODE-based samplers, necessitating ad hoc modifications \citep{liu2025flow,xue2025dancegrpo}. In contrast, proximal diffusion provides an efficient yet stochastic SDE-based sampler with speed comparable to ODE methods \citep{fang2025beyond}, making it naturally amenable to RL. Yet, the distinct non-Gaussian transition kernel in proximal-based samplers introduces challenges for applying RL objectives, which are defined using transition densities. This motivates the main question in this work: \emph{Can we combine the improved speed of proximal diffusion models with the flexibility of reinforcement learning to optimize task-specific rewards in T2I generation?}

We answer this question positively by introducing \textsc{ProxT2I}, a text-conditional proximal diffusion model augmented with reinforcement learning for T2I generation. More specifically, our contributions are:
\begin{enumerate}
    \item We leverage the proximal diffusion framework \citep{fang2025beyond} to develop a new conditional generative model for efficient, high-quality human text-to-image synthesis.
    \item We integrate reinforcement learning via Group Relative Policy Optimization (GRPO) \citep{shao2024deepseekmath} into proximal diffusion, improving perceptual quality and text-image alignment while preserving the fast sampling advantages of proximal-based samplers.
    \item We curate and opensource \texttt{LAION-Face-T2I-15M}, a new large-scale dataset of 15M high-quality human images with fine-grained captions and a 3M hand-focused subset, establishing a new foundation for developing human T2I models. 
    \item  As will be demonstrated, \textsc{ProxT2I} offers a lightweight and efficient solution for fine-grained, text-conditional human image generation with state-of-the-art performance.
    \end{enumerate}

The rest of the paper is organized as follows. \cref{sec:related-works} introduces related works on diffusion models, text-to-image generation, and reinforcement learning. \cref{sec:method} then presents the proposed approach, while our experiments and results are presented in \cref{sec:experiments}. Discussion of limitations and conclusion are finally addressed in	 \cref{sec:conclusion}.

\section{Related Works}
\label{sec:related-works}
\paragraph{Diffusion Models.} Diffusion models have emerged as one of the most popular frameworks for generative modeling, offering a principled way to learn and sample from complex data distributions through iterative denoising \citep{sohl2015deep,ho2020denoising,song2021score}. Diffusion models sample data by simulating the reverse of a forward noising process \citep{song2021score}, where the reverse-time dynamics are governed by an SDE (or its deterministic ODE counterpart). Sampling then amounts to discretizing this reverse process, typically using explicit solvers analogous to Euler–Maruyama schemes, resulting in updates that invoke the gradient of the log-density (\emph{a.k.a.} the score) at each step \citep{song2021score}. Compared to conventional approaches such as VAEs or GANs, diffusion models are favored for their training stability, high sample quality, and natural flexibility to conditional generation through mechanisms like classifier-free guidance \citep{dhariwal2021diffusion,ho2022classifier}. However, these advantages come at the cost of computational inefficiency, as explicit, forward discretizations of the reverse process typically require a large number of small sampling steps to ensure good sample fidelity, and score-based updates can be sensitive to hyperparameter choice and lack of regularity in the data distribution \citep{karras2022elucidating,chen2023improved,beyler2025optimal}. \emph{Proximal Diffusion Models (ProxDM)} \citep{fang2025beyond} offer an alternative formulation by applying a \emph{backward discretization} to the reverse SDE, casting each update as an implicit step that defines a proximal operator of the log-density of the data distribution (see more details in \cref{sec:method}). This enhances stability and allows considerably larger step sizes and faster convergence, providing a principled and efficient foundation for generative sampling.

\paragraph{Diffusion models and reinforcement learning for T2I generation.} 
Diffusion models have achieved remarkable success in text-to-image generation with power approaches such as Stable Diffusion \citep{rombach2022ldm}, Imagen \citep{saharia2022imagen}, and DALL$\cdot$E~2 \citep{ramesh2022hierarchical}. However, as with unconstrained sampling, these methods require many sampling steps to achieve high quality. Considerable work has focused on accelerating inference through ODE-based samplers and flow-matching models \citep{song2021denoising,lipman2022flow,lu2022dpm}, or via knowledge distillation \citep{salimans2022progressive,luhman2021knowledge,chen2025denoising}. Despite these advances, few-step samplers often require additional distillation training or suffer from degraded fidelity or limited robustness.
Beyond conventional pretraining on data samples, reinforcement learning (RL) has been explored to directly align generative models with downstream objectives or human preferences \citep{black2024training,liu2025flow,xue2025dancegrpo,hu2025towards}. GRPO-based approaches in particular improve stability and scalability but often rely on modified versions of deterministic flow-based samplers with added noise for stochastic exploration \citep{liu2025flow,xue2025dancegrpo}. Our work will instead leverage the intrinsic stochasticity of proximal diffusion, enabling a natural integration with GRPO for efficient, reward-guided text-to-image generation.

\section{Methods}
\label{sec:method}
We now first introduce some background on diffusion models, laying the necessary elements for \textsc{ProxT2I}. We will then introduce our conditional proximal diffusion model, and demonstrate how a proximal variant of classifier-free guidance \citep{ho2021classifier} can be applied to flexibly control the conditioning strength and improve image quality. Finally, we will describe how RL, and the GRPO algorithm in particular, can be incorporated with proximal sampling for task-specific reward optimization.

\subsection{Warm-up: Proximal Diffusion Models}

Diffusion models are built upon reversing a stochastic forward process that progressively perturbs data toward a simple and known distribution, typically a standard Gaussian.
A common choice for this is the Ornstein–Uhlenbeck (OU) process, described by the following SDE:
\begin{align*}
dX_t = -\tfrac{1}{2} X_t dt + dW_t, \quad t \in [0,T],
\end{align*}
where $X_t \in \bR^d$ denotes the sample at time $t$ and $W_t \in \bR^d$ is a standard Wiener process (i.e., Brownian motion). Here, $X_0 \sim p_\text{data}$ follows the data (target) distribution. It can be shown \citep{oksendal2013stochastic} that the marginal distribution of $X_t$ is a mixture of the data distribution and a standard normal with time-evolving weights: $X_t \stackrel{d}{=} \sqrt{e^{-t}} X_0 + \sqrt{1 - e^{-t}} \xi, \ \xi \sim \cN(0, I)$. As $T \to \infty$, the distribution of $X_T$ converges to the standard Gaussian. 
In practice, it is common to introduce a scheduling function $\beta(t) \colon \bR \to \bR^+$ to control the diffusion rate while fixing $T=1$:
\begin{align}
\label{eq:forward-process}
dX_t = -\tfrac{1}{2}\beta(t) X_t dt + \sqrt{\beta(t)} dW_t, \quad t \in [0,1].
\end{align}
The corresponding marginal distribution satisfies $X_t \stackrel{d}{=} \sqrt{\alpha_t} X_0 + \sqrt{1 - \alpha_t} \xi,\ \xi \sim \cN(0, I)$, where $\alpha_t = \exp(- \int_0^t \beta(s) ds)$. 
A commonly used choice is a linear schedule: $\beta(t) = \beta_{min} + (\beta_{max} - \beta_{min})t$.

The generative process reverses the forward one by simulating the reverse-time SDE \citep{anderson1982reverse,song2021score}:
\begin{align*}
dX_t = \Bigl[-\tfrac{1}{2}\beta(t)X_t - \beta(t)\nabla \ln p_t(X_t)\Bigr]\,dt + \sqrt{\beta(t)}\,d\bar{W}_t,
\end{align*}
where \(\bar{W}_t\) denotes a reverse-time Wiener process. This SDE preserves the marginal distributions of \eqref{eq:forward-process} \citep{anderson1982reverse}, and thus one can approximate samples from $p_\text{data}$ by drawing $X_{t=1} \sim \cN(0,I)$ and then numerically integrating the reverser-time SDE from $t=1$ to $t=0$.

To make this approach practical, the reverse-time SDE must be discretized and integrated using numerical solvers. A standard choice is the forward discretization, i.e., the explicit Euler--Maruyama scheme:
\begin{align}
X_{k-1} = X_k + \gamma_k \bigl[\tfrac{1}{2} X_k + \nabla \ln p_{t_k}(X_k)\bigr] + \sqrt{\gamma_k}\,\xi_k,
\end{align}
where $X_k$ and $X_{k-1}$ represent the current and next iterates, respectively, $\xi_k \sim \mathcal{N}(0, I)$ and $\gamma_k = \beta(t_k) (t_{k-1} - t_k)$ is the step size scaled by $\beta(t)$. 
In practice, the true score function $\nabla \ln p_t(X_t)$ is approximated by a trained neural network $s_\theta(X_t, t)$, resulting in score-based diffusion models.
This explicit discretization scheme underlies the vast majority of diffusion samplers, including standard \citep{ho2020denoising,song2021score} and accelerated ones \citep{song2021denoising,zhang2023fast,lu2022dpm,wu2024stochastic,lu2025dpm}. 

\paragraph{Proximal Diffusion Models.}
An alternative approach to score-based solvers is to apply \emph{backward discretization}:
\begin{align*}
\label{eq:backward-discretization}
    X_{k-1} = X_k + \gamma_k \bigl[ \tfrac{1}{2} {\color{black} X_{k-1}} + \nabla \ln p_{t_{k-1}}({ \color{black} X_{k-1} })\bigr] + \sqrt{\gamma_k}\,\xi_k,
\end{align*}
where the drift term is evaluated at the next iterate (note the dependence on $X_{k-1}$ on the right-hand side).
Although the update rule for the next iterate $X_{k-1}$ now becomes implicit, this implicit equation can be expressed explicitly using a \emph{proximal operator} \citep{fang2025beyond}. For a scalar-valued (and weakly-convex) function $f \colon \bR^d \to \bR$, its proximal operator is a mapping from $\bR^d$ to $\bR^d$ defined as
\begin{align*}
    \prox_{f}(x) = \arg\min_u f(u) + \tfrac{1}{2} \|u - x\|_2^2,
\end{align*}
which intuitively finds a point $u$ that approximately minimizes $f$ while remaining close to $x$. To see its connection with backward discretization, we can apply the first-order optimality conditions\footnote{We assume $f$ to be smooth for simplicity of presentation, though this is not necessary.} and obtain the following property:
\begin{align}
    \nabla f(\prox_{f}(x)) + \prox_{f}(x) - x = 0,
\end{align}
indicating that $y=\prox_{f}(x)$ is a solution to the implicit equation $\nabla f(y) + y - x = 0$. Thus, by re-arranging the backward-discretized update, one can obtain the \emph{proximal diffusion algorithm} (PDA) from \citep{fang2025beyond},
\begin{align*}
X_{k-1} = \operatorname{prox}_{\tfrac{2\gamma_k}{2-\gamma_k}\,\ln p_{t_{k-1}}}
\!\left(\tfrac{2}{2-\gamma_k}\,(X_k + \sqrt{\gamma_k}\,\xi_k)\right).
\end{align*}
However, this algorithm requires $\gamma_k < 2$ to maintain positive coefficients, imposing a hard constraint on the maximum step size possible.
This restriction can be relaxed using the alternative \emph{hybrid scheme}, which incorporates both forward and backward discretizations:
\begin{align*}
    X_{k-1} = X_k + \gamma_k \bigl[ \tfrac{1}{2} X_{k} + \nabla \ln p_{t_{k-1}}({ \color{black} X_{k-1} })\bigr] + \sqrt{\gamma_k}\,\xi_k,
\end{align*}
yielding the \emph{hybrid proximal diffusion algorithm} (PDA-hybrid):
\begin{align}
X_{k-1} = \operatorname{prox}_{-\gamma_k \ln p_{t_{k-1}}}
\!\Bigl[\,(1+\tfrac{1}{2}\gamma_k) X_k + \sqrt{\gamma_k}\,\xi_k \Bigr].
\end{align}
This formulation lifts the step-size restriction and remains valid at large $\gamma_k$. 

Importantly, the proximal operators in these algorithms can be learned directly from data via proximal matching \citep{fang2024whats}, analogous to score matching for score-based models, thereby enabling \emph{generative modeling}. These proximal diffusion models exhibit theoretically improved iteration complexity compared to forward-discretized, score-based samplers \citep{chen2023improved,benton2024nearly} and achieve empirically faster inference in generating high-quality images \citep{fang2025beyond}. Moreover, because proximal diffusion models rely on SDE discretization, they intrinsically incorporate stochasticity in the sampling trajectory (through the terms $\xi_k$). As we shall see shortly, this property makes PDA and PDA-hybrid particularly amenable to reinforcement learning, which relies on stochasticity in the sampling process for exploration (which is not satisfied by the sampling paths of probability-flow ODEs \citep{song2021denoising,song2021score,zhang2023fast,lu2022dpm} or flow matching models \citep{lipman2022flow,liu2023flow}).

In the following sections, we will extend PDA(-hybrid) to conditional generation and show how to integrate them with GRPO to optimize for task-specific reward models.

\subsection{Proximal Diffusion for Conditional Generation}

Conditional generation is concerned with drawing samples from the conditional distribution of $p_\text{data}(\cdot \mid c)$, where $c \in \bR^{d_c}$ is a conditioning signal, e.g. the embedding of some input text in T2I generation.
The forward process for conditional sampling is defined analogously to the unconditional case \eqref{eq:forward-process}, but initialized at the conditional data distribution: $X_0 \sim p_\text{data}(\cdot \mid c)$. This conditional forward process admits the following reverse-time SDE:
\begin{align}
\label{eq:conditional-reverse-SDE}
dX_t = \bigl[-\tfrac{1}{2}\beta(t)X_t - \beta(t)\nabla \ln p_t(X_t \mid c)\bigr]dt
       + \sqrt{\beta(t)}\,d\bar{W}_t,
\end{align}
where $\nabla \ln p_t(\cdot \mid c)$ is the score of the data distribution of the conditional forward process at time $t$. To sample from $p_\text{data}(\cdot \mid c)$, score-based samplers discretize \cref{eq:conditional-reverse-SDE} using a forward Euler scheme while parameterizing the conditional score at each step with a neural network that takes $c$ as an additional input.
In contrast, in this work we leverage the hybrid discretization from PDA-hybrid to integrate \cref{eq:conditional-reverse-SDE}:
\begin{align*}
    X_{k-1} = X_k + \tfrac{1}{2}\gamma_k \bigl[X_{k} + \nabla \ln p_{t_{k-1}}( X_{k-1} \mid c)\bigr] + \sqrt{\gamma_k}\,\xi_k,
\end{align*}
which can be rewritten as:
\begin{align}
\label{eq:conditional-proxdm}
X_{k-1}
= \prox_{-\gamma_k \ln p_{t_{k-1}}(\cdot \mid c)}
\!\Bigl[(1+\tfrac{1}{2}\gamma_k)X_k + \sqrt{\gamma_k}\,\xi_k\Bigr],
\end{align}
by the first-order optimality condition of the proximal.
Each iteration thus evaluates the proximal operator of the \emph{conditional} log-density $\ln p_{t_{k-1}} (\cdot \mid c)$, guiding samples toward high-likelihood regions of the conditional data distribution.

Although the vanilla algorithm in \cref{eq:conditional-proxdm} can in principle achieve conditional generation, in practice it often produces samples with suboptimal quality and weak coherence to the conditioning signal, as also observed in score-based models \citep{ho2021classifier,rombach2022ldm}. A widely adopted solution in popular text-to-image models \citep{nichol2021glide,rombach2022ldm,podell2024sdxl,esser2024scaling} is \emph{classifier-free guidance} (CFG) \citep{ho2021classifier,ho2022classifier}, which strengthens the conditional signal by linearly combining the conditional and unconditional scores.
Inspired by this approach, we introduce a proximal variant of the CFG technique:
\begin{align}
\label{eq:proxdm-cfg-step}
X_{k-1}
&= f_\omega
\left[(1+\tfrac{1}{2}\gamma_k)X_k + \sqrt{\gamma_k}\,\xi_k ; c \right] 
\end{align}
by using the linear combination of conditional and unconditional proximal mappings:
\begin{align*}
    f_\omega(x ; c) = ( 1+\omega ) \prox_{-\gamma_k \ln p_{t_{k-1}}(\cdot \mid c)} (x) \\
    - \omega \prox_{-\gamma_k \ln p_{t_{k-1}}} (x).
\end{align*}
When $\omega = 0$, this reduces to the standard conditional proximal update in \cref{eq:conditional-proxdm}, while for $\omega > 0$, the update direction is amplified by the discrepancy between the conditional and unconditional proximal mappings--thereby steering each step toward regions favored by the conditioning signal and away from the unconditional samples.
Analogous to score models, larger $\omega$ values encourage stronger semantic alignment with the conditioning input $c$. 
As we will demonstrate in our experiments, this guidance mechanism substantially enhances the sample fidelity and conditioning consistency of the proximal diffusion sampler while maintaining its high sampling efficiency.

\paragraph{Learning Conditional Proximal Functions.}
To achieve generative modeling using the sampling scheme above, the time-specific proximal operators of the log-density must be learned from data.
We consider a neural network $f_\theta(x; t, \lambda, c)$ that parameterizes the conditional proximal operator $\prox_{- \lambda \ln p_{t}(\cdot \mid c)} (x)$, and our goal is to train $f_\theta$ for all possible $\{t, \lambda, c\}$ tuples.
Similar to the unconditional proximal models, we train these conditional proximals using the \emph{proximal matching} objective from \citet{fang2024whats}, which resembles denoising score matching but replaces the mean squared error with a specialized loss that encourages approximation of the mode of a conditional distribution--coinciding with the interpretation of proximal operators as the Maximum A Posteriori (MAP) denoiser under Gaussian noise.
To train these operators, we minimize the following loss:
\begin{align}
\label{eq:true-risk-prox-matching-loss}
    \cL(\theta) = \underset{t, \lambda, c, X_t, \varepsilon}{\bE} \left\{ \ell_{\mathrm{PM}} \left[ f_\theta( X_t + \sqrt{\lambda} \varepsilon; t, \lambda, c ) , X_t; \zeta \right] \right\}
\end{align}
where $(t, \lambda)$ is drawn from a predefined distribution $p(t,\lambda)$, $X_t \sim p_t(\cdot \mid c)$ is sampled from the forward process at time $t$ (conditioned on $c$), and $\varepsilon \sim \cN(0, I)$ is standard Gaussian noise. Importantly,
\begin{align*}
    \ell_{\mathrm{PM}} (x, y; \zeta) = 1 - \exp\left( -\frac{\|x - y\|_2^2}{d \zeta^2} \right)
\end{align*}
is the proximal matching loss \citep{fang2024whats}, where
$\zeta \in \bR^+$ is a hyperparameter that balances the approximation error to the proximal and the smoothness of the loss function.
It can be shown that the minimizer of \cref{eq:true-risk-prox-matching-loss},  $f_{\theta^*}(x; t, \lambda, c)$, approaches the desired proximal operator, $\prox_{-\lambda \ln p_t(\cdot \mid c)}(x)$, as $\zeta \to 0$. 

To support the CFG technique, we will also need to learn the unconditional proximal operator $\prox_{-\lambda \ln p_t}(x)$. Following standard practice \citep{ho2022classifier}, we parameterize both the unconditional and conditional proximal functions using a single neural network, and set the input condition $c$ to a null context $\varnothing$ for evaluating the unconditional model. In text-to-image generation, this corresponds to using the embedding of an empty text as $c$. To train the unconditional model, we simply replace $c$ with $\varnothing$ in each training sample $(X_t, c)$ following a predefined probability $p_{\varnothing}$ (we let $p_{\varnothing} = 0.1$ following prior work \citep{bao2023all}). This design effectively enables the network to learn the proximal operators of both the conditional and unconditional log-densities.

\subsection{Proximal Diffusion Policy Optimization}
The conditional proximal diffusion described above already enables sampling from the desired conditional distribution. Indeed, in the idealized limit of infinite training data (and compute), infinite network capacity, and an unlimited budget of sampling steps, it would generate samples that match the data distribution. However, beyond faithfully matching a data distribution, many use cases of generative models are more concerned with task-specific, downstream objectives, such as human-perceived quality and prompt-image alignment \citep{black2024training}. We will now show how to apply reinforcement learning to directly optimize our proximal models for such objectives. 
Consider a Markov decision process (MDP) whose rounds are indexed by $k$, with state $s_k$, action $a_k$, reward $R(s_k,a_k)$, and policy $\pi_\theta$. The goal of reinforcement learning is to learn a policy that maximizes the expected time-cumulative reward over sampled trajectories:
\if\isArxiv 1
\begin{align*}
\max_\theta\;
\mathbb{E}_{(s_0,a_0,\dots,s_K,a_K)\sim \pi_\theta}
\!\left[\;
\sum_{k=0}^{K} R(s_k,a_k)
\right].
\end{align*}
\fi
\if\isArxiv 0
$\max_\theta\;
\mathbb{E}_{(s_0,a_0,\dots,s_K,a_K)\sim \pi_\theta}
\!\left[\;
\sum_{k=0}^{K} R(s_k,a_k)
\right].$
\fi
To apply these ideas to diffusion models, the sampling procedure is typically mapped to an MDP \citep{black2024training}, where the state is a tuple of conditioning, step index, and the current sample: $s_k \triangleq (c, k, X_k)$, the action is the next sample $a_k \triangleq X_{k-1}$, and the policy is the corresponding transition kernel $\pi_\theta(a_k \mid s_k) \triangleq p_\theta(X_{k-1} \mid X_k, c)$, describing how the sample evolves from step $k$ to $k-1$. 
To optimize this policy, %
we adopt Group Relative Policy Optimization (GRPO) \citep{shao2024deepseekmath}, a lightweight and scalable variant of Proximal Policy Optimization (PPO) \citep{schulman2017proximal} that has been successfully applied to text-to-image flow models \citep{liu2025flow}. For each conditioning $c$, GRPO first samples a group of trajectories $\{\tau^{(i)}\}_{i=1}^{G}$, with $ \tau^{(i)} = \left( X_{K}^{(i)}, X_{K-1}^{(i)}, \dots, X_0^{(i)} \right)$. The final outputs $\{X_0^{(i)}\}$ are evaluated by a reward function $R(X_0^{(i)}, c)$, and the \emph{group-relative advantages} are then computed by:
\begin{align*}
    \hat{A}^{(i)} = \frac{R(X_0^{(i)}, c) - \operatorname{mean} ( \{ R(X_0^{(i)}, c) \}_{i=1}^G ) }{ \operatorname{std} ( \{ R(X_0^{(i)}, c) \}_{i=1}^G ) }.
\end{align*}
Such a formulation aligns with the comparative nature of most reward models and provides a more stable and informative signal for policy optimization \citep{shao2024deepseekmath}. To use these computed advantage to perform policy optimization, we can maximize the following objective\footnote{Here we present a simplified version of the GRPO objective for simplicity of presentation. The full definition can be found in \cref{sec:full-grpo-objective}.}:
\begin{align*}
&\mathcal{J}_{\mathrm{GRPO}}(\theta) = 
 \sum_{i=1}^G \sum_{k=1}^K 
\left\{ 
R_k^{(i)} (\theta) - \beta_{\mathrm{KL}} D_{\mathrm{KL},k}^{(i)}(\theta)
\right\} ,
\end{align*}
where
\begin{align*}
\label{eq:clipped-reward}
R_k^{(i)} (\theta) = \frac{p_\theta (X^{(i)}_{k-1} \mid X^{(i)}_{k}, c)}{p_{\theta_{\text{old}}}(X^{(i)}_{k-1} \mid X^{(i)}_{k}, c)} \hat{A}^{(i)},
\end{align*}
with $\theta_{\text{old}}$ denoting the parameters before the current update step. Intuitively, the objective encourages the model to assign higher probability to the samples $X^{(i)}_{k-1}$ from paths with positive advantages, while penalizing those with negative ones.
The KL regularization term
\begin{align*}
D_{\mathrm{KL},k}^{(i)}(\theta) = \KL \left[ p_\theta( \cdot \mid X^{(i)}_{k}) \ \| \  p_{\theta_\text{ref}} (\cdot \mid X^{(i)}_{k}) \right]
\end{align*}
prevents the policy from drifting too far from a fixed reference model $\theta_\text{ref}$, typically set as the pretrained model before reinforcement learning, and $\beta_{\mathrm{KL}} \in \bR^+$ controls the strength of regularization.
As the objective shows, optimizing $\theta$ hinges on evaluation of the transition kernel $ p_\theta( X^{(i)}_{k-1} \cdot \mid X^{(i)}_{k}, c) $ and its gradient with respect to $\theta$ -- which we derive next.

\paragraph{Integrating RL with proximal diffusion.}
For score-based diffusion models, the transition kernel $p_\theta(X_{k-1} \mid X_{k}, c)$ is Gaussian \citep{black2024training} and permits efficient evaluation of the GRPO objective. 
Nonetheless, the transition kernel corresponding to the proximal diffusion update (\cref{eq:proxdm-cfg-step}) is:
\begin{align*}
p_\theta(X_{k-1} \mid X_k, c) = \left[f_\omega(\cdot ; c)\right]_{\#} \mathcal{N}( (1+\tfrac{1}{2}\gamma_k) X_k, \gamma_k I),
\end{align*}
where $h_\#$ denotes the pushforward of a distribution under the mapping $h$. Equivalently, $X_{k-1}$ is given by pushing forward a Gaussian through the deterministic mapping $f_\omega(\cdot ; c)$, a linear combination of conditional and unconditional proximal networks. This results in a \emph{non-Gaussian} transition kernel that is, in general, intractable to evaluate. Only under strong assumptions, such as when $f_\omega$ is invertible, one can leverage change-of-variables to obtain an analytical expression for the ratio of kernels.
However, since the proximal networks in our formulation are not constrained to be bijective or have tractable Jacobian determinants, direct computation of $p_\theta$ remains infeasible.

To address this issue, we introduce an auxiliary variable that yields tractable Gaussian transitions. Let
\begin{align}
Y_k &= (1 + \tfrac{1}{2}\gamma_k) X_k + \sqrt{\gamma_k}\,\xi_k, 
\quad k = 1, \dots, K.
\end{align}
and $Y_0 = X_0$.
Then, the sampling process in \eqref{eq:proxdm-cfg-step} can be rewritten as
\begin{align*}
&Y_K  \sim \mathcal{N}(0, [(1+\tfrac{1}{2}\gamma_K)^2 + \gamma_K] I), \\
&Y_{k-1} = (1 + \tfrac{1}{2}\gamma_{k-1}) f_\omega(Y_k; c) + \sqrt{\gamma_{k-1}} \xi_{k-1}, \; k = K, \dots, 2, \\
&Y_0 = f_\omega(Y_1; c),
\end{align*}
and the resulting variables $\{Y_k\}_{k=0}^K$ now admit explicit Gaussian transition kernels:
\begin{align*}
&p_\theta(Y_{k-1} \mid Y_k, c) = \mathcal N \left((1+\tfrac{1}{2}\gamma_{k-1})\,f_\omega(Y_k; c),\ \gamma_{k-1} I \right),
\end{align*}
for $k = K, \dots, 2$\footnote{We use $\mathcal N$ to denote the density of a Gaussian.}.
Under this new formulation, the GRPO objective can be directly applied to the proximal sampler through the MDP defined by $\{Y_k\}_{k=0}^{K}$, enabling tractable evaluation of the GRPO objective in the proximal setting. 

The application of RL to generative models based on efficient ODE and flow-based samplers often requires careful modifications of the sampling process to inject stochasticity necessary for exploration \citep{liu2025flow,xue2025dancegrpo}. 
In contrast, proximal diffusion models provide comparable sampling efficiency to ODE-based samplers while retaining the stochasticity from SDE discretization, allowing direct application of RL without modification of the sampling path. Notably, the efficiency and stochasticity of proximal sampling could also be useful in other methods that require random transitions, such as inference-time reward alignment \citep{holderrieth2025glass}.

\subsection{LAION-Face-T2I-15M Dataset}

Although text-to-image generation models have achieved remarkable progress, finegrained generation of human features (such as hands and faces) remains a major challenge. While methods keep evolving, existing open datasets are either limited in scale, domain-specific (e.g., face-only), or lack comprehensive and rich textual descriptions \citep{xia2021tedigan,xia2021towards,ju2023human,li2024cosmicman}. Thus, in this work we also curate, collect, and opensource \textsc{LAION-Face-T2I-15M}, a dataset  of $15$M text-image pairs designed for human-domain text-to-image generation.
LAION-Face-T2I-15M builds upon LAION-Face\footnote{\url{https://github.com/FacePerceiver/LAION-Face}}, a filtered subset of LAION-400M curated to contain high-quality images with human faces.
We further filter the dataset based on image size and aesthetic score \citep{schuhmann2022laionaesthetics}, resulting in a high-resolution and high-quality collection of roughly $15$M images. To improve the details in caption for fine-grained generation, we recaption all images using the Qwen-2.5-7B vision-language model \citep{qwen2025qwen25technicalreport} with a carefully-designed prompt that promotes accurate, comprehensive, and fine-grained textual descriptions.  The resulting captions describe not only facial attributes but also hair, makeup, clothing, ethnicity\footnote{Automatically generated captions may be biased or inaccurate in describing perceived ethnicity or gender, and large-scale human datasets can inherit demographic biases from their data sources. Unfortunately, the LAION dataset (from which we derive ours) does not have ground truth demographic attributes, as they are images crawled from the web, and thus no real analysis can be provided. We refer the reader to the recent discussion in \citep{girrbach2025person} for the LAION dataset, and advise caution when employing any demographic attributes in our dataset.},
image background and overall image style (e.g., photography vs. artwork). This process yields high-quality captions suitable for training human text-to-image models from scratch. Additionally, we also construct LAION-Face-Hand-3M, a subset of LAION-Face-T2I-15M tailored for hand image generation. We employ Qwen-2.5-7B with a dedicated prompt to identify images containing close-up views of hands and generate detailed captions describing the hand attributes. This results in approximately 3M images containing human hands paired with descriptive captions that describe the number of hands, their position relative to the body, finger gestures, interactions (e.g., holding objects or touching the face), and visibility quality. This provides valuable data for learning human-hand generation, an area where current models often fail.

We will publicly release LAION-Face-T2I-15M and LAION-Face-Hand-3M, including all generated captions and image URLs, upon publication of this paper. By combining careful filtering, high-quality recaptioning, and targeted hand-focused subsets, we hope this collection of data will serve as a valuable resource for the research community and facilitate the pretraining and finetuning of human-domain text-to-image models. More details on data processing, including detailed prompts, can be found in \cref{sec:appx-t2i-dataset}. Examples from the dataset are provided in \cref{fig:dataset-examples}.

\section{Experiments and Results}

\label{sec:experiments}
We perform generation in the latent space of a variational autoencoder (VAE) \citep{rombach2022ldm,esser2024scaling} and use a combination of text encoders (MetaCLIP \citep{xu2023demystifying} and T5 \citep{raffel2020exploring}) for textual embedding. The U-ViT network \citep{bao2023all} is adopted as the backbone for both proximal and competing score networks for its flexibility in handling varying input dimensions. All models are trained from scratch on the LAION-Face-T2I-15M dataset and subsequently finetuned with GRPO. More experimental details can be found in \cref{sec:appx-experimental-details}.

We compare against multiple score-based samplers: 1) ``SDE-Euler'' employs the standard Euler-Maruyama discretization on the reverse-time SDE in \cref{eq:conditional-reverse-SDE}; 2) ``ODE-Euler'' applies Euler discretization to the probability flow ODE \citep{song2021score}; and 3) ``DPM-solver'' is the fast higher-order solver proposed in \citep{lu2022dpm}. Note that only the SDE-Euler is a ``fair'' comparison model (since ProxT2I also discretizes an SDE) but, as we will see, ProxT2I also outperforms ODE and DPM solvers. We use CFG with $\omega=4$ for all the models. To facilitate quantitative comparison, we compute four metrics estimating human preference alignment and aesthetic quality: HPSv2.1 \citep{wu2023human}, Aesthetic score \citep{schuhmann2022laionaesthetics}, ImageReward \citep{xu2023imagereward}, and PickScore \citep{kirstain2023pick}. 

\begin{figure}[t]
    \centering
    \if\isArxiv 1
    \includegraphics[width=0.6\textwidth,trim=0 0 0 325,clip]{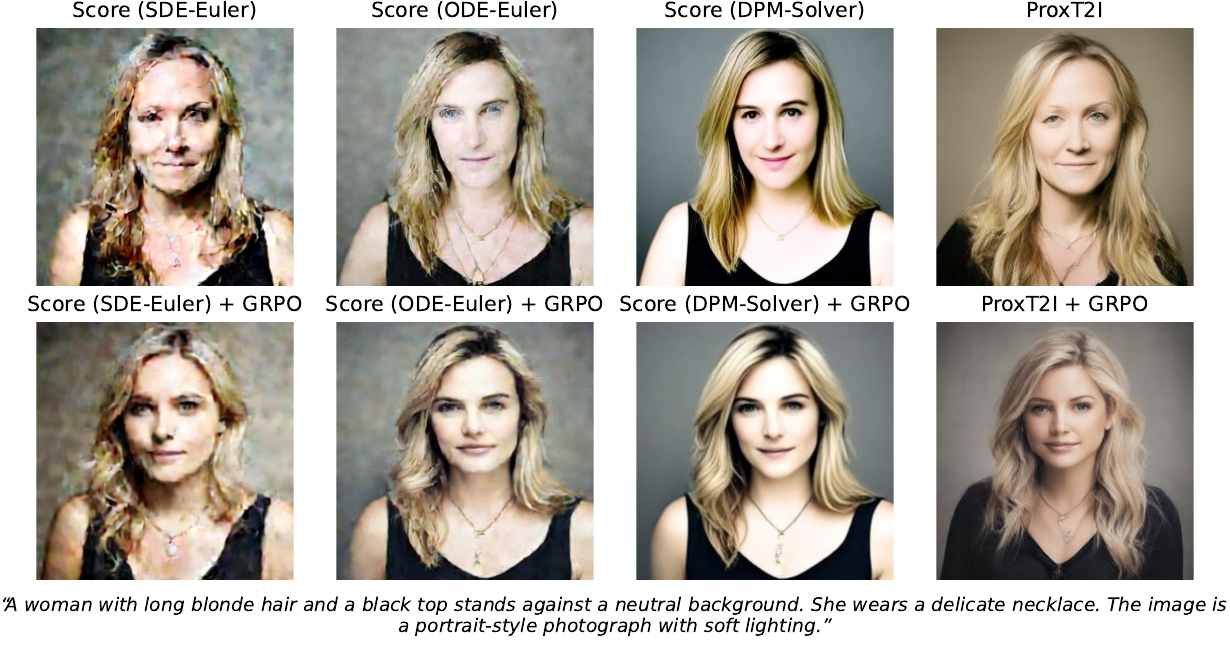}
    \vspace{-10pt}
    \fi
    \if\isArxiv 0
    \includegraphics[width=0.5\textwidth,trim=0 0 0 325,clip]{figures/comparison_grid_256_10steps.pdf}
    \vspace{-20pt}
    \fi
    \caption{Samples generated by ProxT2I and competing methods at $256^2$ resolution using 10 sampling steps.}
    \label{fig:256-main-samples}
\end{figure}

\subsection{Comparison with competing methods}

\paragraph{Results at $256^2$ Resolution.}
\cref{fig:256} presents the HPSv2.1 scores at $256^2$ resolution across different samplers and step counts and \cref{fig:256-main-samples} visualizes image samples. \cref{fig:256-images} in Appendix presents additional samples and \cref{fig:256-metric-appx} provides additional metrics. Results are averaged over a held-out testing set of 1024 prompts. Among non-RL methods, ProxT2I already outperforms SDE-Euler and ODE-Euler and is competitive with DPM-Solver. When fine-tuned with GRPO, the numerical results of ProxT2I significantly improves, demonstrating the effectiveness of the proposed proximal diffusion policy optimization approach. Among all methods, the improvements from GRPO are most pronounced for ProxT2I. ProxT2I--GRPO consistently outperforms score-based RL-finetuned models and achieves the highest performance across sampling step numbers and metrics, demonstrating efficient and human-preference-aligned T2I generation. Notably, ProxT2I--GRPO also achieves strong results at just 4 sampling steps, despite the fact that RL training was applied to the Markov decision process at 10 sampling steps, demonstrating good generalization ability of RL and sharper and more human-preference-aligned generations even in very low step regimes. %

\begin{figure*}[t]
    \centering
    \subcaptionbox{HPSv2.1 score.}{%
        \includegraphics[width=0.44\textwidth]{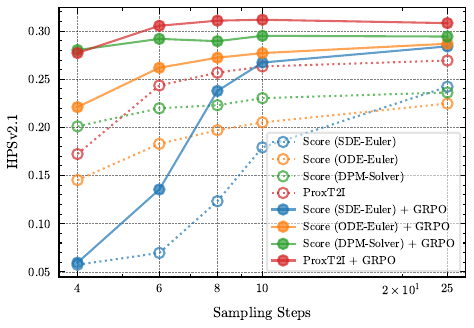}}
    \subcaptionbox{Images generated using 10 sampling steps.}{%
        \includegraphics[width=0.55\textwidth,trim=0 0 0 325,clip]{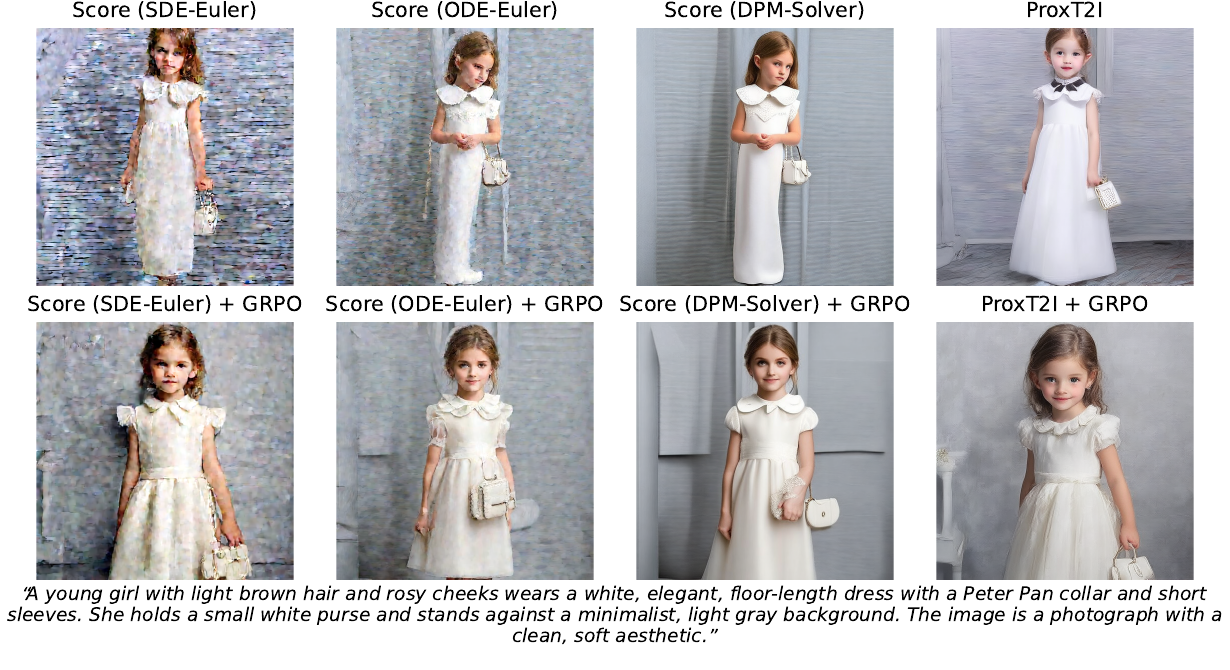}}

    \caption{Comparison between ProxT2I and competing methods at $512^2$ resolution.}
    \label{fig:512}
\end{figure*}

\paragraph{Results at $512^2$ Resolution.}
We further evaluate the models' performance for high-resolution generation at $512^2$. \cref{fig:512} presents HPSv2.1 scores and visual samples (additional metrics and samples can be found in \cref{fig:512-metric-appx,fig:512-images}, respectively). Notably, ProxT2I achieves notable improvements over all socre-based solvers both before and after RL finetuning, particularly in low-step sampling regimes, demonstrating that the efficiency benefits of ProxT2I scale to higher-dimensions.

\paragraph{Comparison with external open-source models.}

\begin{figure}[t]
    \centering
    \subcaptionbox{HPSv2.1 score at $256^2$ resolution.}{%
        \includegraphics[width=0.39\textwidth]{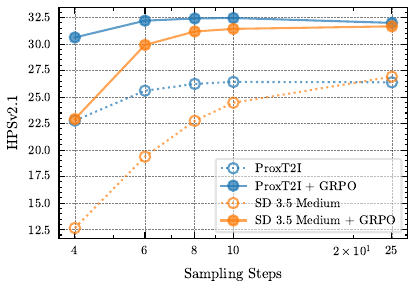}}
    \subcaptionbox{HPSv2.1 score at $512^2$ resolution.}{%
        \includegraphics[width=0.39\textwidth]{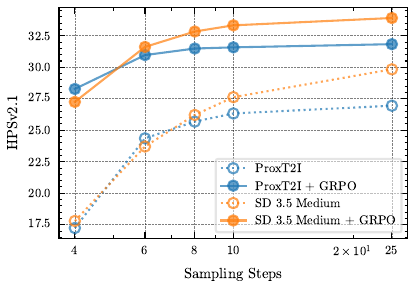}}
    \caption{Comparison with Stable Diffusion (SD) 3.5 Medium \citep{esser2024scaling} and its Flow-GRPO-finetuned variant \citep{liu2025flow}.}
    \vspace{-5pt}
    \label{fig:external}
\end{figure}

To evaluate our model's performance among widely-used text-to-image models, we compare with Stable Diffusion (SD) 3.5 Medium \citep{esser2024scaling} and its respective GRPO-fine-tuned variant \citep{liu2025flow}, with HPSv2.1 results shown in \Cref{fig:external} and additional metrics in \cref{fig:external-appx}.
It is important to note that this comparison is not strictly fair: ProxT2I employs a lighter backbone (a transformer with approximately 500 million parameters), 
and is trained using data and protocols that will be publicly released. In contrast, SD 3.5 uses a substantially larger transformer with about 2.2 \emph{billion} parameters 
and is trained on a mixture of synthetic and filtered public data, though the specific details on training strategy and dataset are not fully disclosed. Furthermore, SD 3.5 is a general-domain model, while ProxT2I is trained specifically for human image synthesis.
Despite these differences, ProxT2I achieves competitive performance relative to SD 3.5, particularly at low sampling steps. These results demonstrate that ProxT2I offers a lightweight and efficient solution for human-domain text-to-image generation that requires substantially less compute and memory without significantly degrading generation quality.

\subsection{Ablation Studies}
\paragraph{Classifier-free guidance.} We further study the influence of the classifier-free guidance (CFG) level, $\omega$, in our results. \cref{fig:cfg} shows the numerical results at different $\omega$ values on $256^2$ images over a validation set of 256 prompts (model trained without GRPO), with visual samples shown in  \cref{fig:cfg-images}. Note that $\omega=0$ represents the sampler without classifier-free guidance, i.e., using only the conditional proximal operator. It is clear that using CFG ($\omega > 0$) effectively improves image quality, with the performance peaking around $\omega=4.0$. Note also that this parameter is quite stable as a function of the number of sampling steps.

\paragraph{Ablation of reward models.}
To study the influence of different reward models in reinforcement learning, we finetune the pretrained ProxT2I model at $256^2$ resolution with two other reward functions--ImageReward \citep{xu2023imagereward} and Aesthetic score \citep{schuhmann2022laionaesthetics}--in addition to PickScore \citep{kirstain2023pick}.
\cref{fig:ablation-reward} demonstrates that training with each reward function leads to a noticeable improvement in its corresponding metric over the pretrained baseline, confirming the effectiveness of RL fine-tuning for the proximal diffusion model.
Among the three, the PickScore reward produces the most consistent improvements across all metrics and achieves the largest gain on the external metric HPSv2.1.
The Aesthetic Score reward, while significantly enhancing the result on its own, leads to degradation in all other metrics due to reward hacking:
as depicted in \cref{fig:aesthetic-collapse}, the model initially learns to generate more aesthetically appealing images but eventually collapses, generating virtually identical images for all input prompts.

\section{Conclusion and Limitations}

\label{sec:conclusion}
We presented ProxT2I(-GRPO), a human-domain text-to-image framework that unifies proximal diffusion models with reinforcement learning. By extending proximal sampling to conditional generation and integrating it with policy optimization, our approach directly addresses the challenges of efficiency, generation quality, and human-preference alignment in few-step T2I generation. We further presented LAION-Face-T2I-15M, a large-scale dataset with fine-grained captions that provides a strong foundation for developing human T2I models. Experiments demonstrate that ProxT2I-GRPO achieves state-of-the-art performance at very low step counts, surpassing score-based samplers and performing competitively with larger open-source systems despite requiring fewer parameters. These results highlight the potential of proximal diffusion models for efficient, high-quality and reward-guided generation.

Our approach has several limitations and potential societal impacts. RL is sensitive to the design of the reward model, which may encode undesired biases from the training data, and optimizing for such signals risks overfitting to spurious correlations rather than true human preference. Human image generation and large-scale human image dataset can also raise ethical concerns related to privacy, consent, and fairness, and models may perpetuate demographic biases in data sources. The lack of ground-truth demographic data in LAION complicates the analysis of this limitation further. Finally, while ProxT2I--GRPO demonstrates strong performance for human-domain T2I, future work will include extending it to broader domains, exploring alternative reward signals for alignment, and scaling to ultra-high resolutions. {
    \small
    \bibliographystyle{ieeenat_fullname}
    \bibliography{refs}

\begin{thebibliography}{59}
\providecommand{\natexlab}[1]{#1}
\providecommand{\url}[1]{\texttt{#1}}
\expandafter\ifx\csname urlstyle\endcsname\relax
  \providecommand{\doi}[1]{doi: #1}\else
  \providecommand{\doi}{doi: \begingroup \urlstyle{rm}\Url}\fi

\bibitem[Anderson(1982)]{anderson1982reverse}
Brian~DO Anderson.
\newblock Reverse-time diffusion equation models.
\newblock \emph{Stochastic Processes and their Applications}, 12\penalty0
  (3):\penalty0 313--326, 1982.

\bibitem[Bao et~al.(2023)Bao, Nie, Xue, Cao, Li, Su, and Zhu]{bao2023all}
Fan Bao, Shen Nie, Kaiwen Xue, Yue Cao, Chongxuan Li, Hang Su, and Jun Zhu.
\newblock All are worth words: A vit backbone for diffusion models.
\newblock In \emph{Proceedings of the IEEE/CVF conference on computer vision
  and pattern recognition}, pages 22669--22679, 2023.

\bibitem[Benton et~al.(2024)Benton, Bortoli, Doucet, and
  Deligiannidis]{benton2024nearly}
Joe Benton, Valentin~De Bortoli, Arnaud Doucet, and George Deligiannidis.
\newblock Nearly \$d\$-linear convergence bounds for diffusion models via
  stochastic localization.
\newblock In \emph{The Twelfth International Conference on Learning
  Representations}, 2024.

\bibitem[Beyler and Bach(2025)]{beyler2025optimal}
Eliot Beyler and Francis Bach.
\newblock Optimal denoising in score-based generative models: The role of data
  regularity.
\newblock \emph{arXiv preprint arXiv:2503.12966}, 2025.

\bibitem[Black et~al.(2024)Black, Janner, Du, Kostrikov, and
  Levine]{black2024training}
Kevin Black, Michael Janner, Yilun Du, Ilya Kostrikov, and Sergey Levine.
\newblock Training diffusion models with reinforcement learning.
\newblock In \emph{The Twelfth International Conference on Learning
  Representations}, 2024.

\bibitem[Chen et~al.(2023)Chen, Lee, and Lu]{chen2023improved}
Hongrui Chen, Holden Lee, and Jianfeng Lu.
\newblock Improved analysis of score-based generative modeling: User-friendly
  bounds under minimal smoothness assumptions.
\newblock In \emph{International Conference on Machine Learning}, pages
  4735--4763. PMLR, 2023.

\bibitem[Chen et~al.(2025)Chen, Zhang, Wang, Wu, Leong, and
  Zhou]{chen2025denoising}
Tianyu Chen, Yasi Zhang, Zhendong Wang, Ying~Nian Wu, Oscar Leong, and Mingyuan
  Zhou.
\newblock Denoising score distillation: From noisy diffusion pretraining to
  one-step high-quality generation.
\newblock \emph{arXiv preprint arXiv:2503.07578}, 2025.

\bibitem[Clark et~al.(2023)Clark, Vicol, Swersky, and Fleet]{clark2023directly}
Kevin Clark, Paul Vicol, Kevin Swersky, and David~J Fleet.
\newblock Directly fine-tuning diffusion models on differentiable rewards.
\newblock \emph{arXiv preprint arXiv:2309.17400}, 2023.

\bibitem[Dhariwal and Nichol(2021)]{dhariwal2021diffusion}
Prafulla Dhariwal and Alexander Nichol.
\newblock Diffusion models beat gans on image synthesis.
\newblock \emph{Advances in neural information processing systems},
  34:\penalty0 8780--8794, 2021.

\bibitem[Dosovitskiy et~al.(2021)Dosovitskiy, Beyer, Kolesnikov, Weissenborn,
  Zhai, Unterthiner, Dehghani, Minderer, Heigold, Gelly, Uszkoreit, and
  Houlsby]{dosovitskiy2021an}
Alexey Dosovitskiy, Lucas Beyer, Alexander Kolesnikov, Dirk Weissenborn,
  Xiaohua Zhai, Thomas Unterthiner, Mostafa Dehghani, Matthias Minderer, Georg
  Heigold, Sylvain Gelly, Jakob Uszkoreit, and Neil Houlsby.
\newblock An image is worth 16x16 words: Transformers for image recognition at
  scale.
\newblock In \emph{International Conference on Learning Representations}, 2021.

\bibitem[Esser et~al.(2024)Esser, Kulal, Blattmann, Entezari, M{\"u}ller,
  Saini, Levi, Lorenz, Sauer, Boesel, Podell, Dockhorn, English, and
  Rombach]{esser2024scaling}
Patrick Esser, Sumith Kulal, Andreas Blattmann, Rahim Entezari, Jonas
  M{\"u}ller, Harry Saini, Yam Levi, Dominik Lorenz, Axel Sauer, Frederic
  Boesel, Dustin Podell, Tim Dockhorn, Zion English, and Robin Rombach.
\newblock Scaling rectified flow transformers for high-resolution image
  synthesis.
\newblock In \emph{Forty-first International Conference on Machine Learning},
  2024.

\bibitem[Fan et~al.(2023)Fan, Watkins, Du, Liu, Ryu, Boutilier, Abbeel,
  Ghavamzadeh, Lee, and Lee]{fan2023dpok}
Ying Fan, Olivia Watkins, Yuqing Du, Hao Liu, Moonkyung Ryu, Craig Boutilier,
  Pieter Abbeel, Mohammad Ghavamzadeh, Kangwook Lee, and Kimin Lee.
\newblock Dpok: Reinforcement learning for fine-tuning text-to-image diffusion
  models.
\newblock \emph{Advances in Neural Information Processing Systems},
  36:\penalty0 79858--79885, 2023.

\bibitem[Fang et~al.(2024)Fang, Buchanan, and Sulam]{fang2024whats}
Zhenghan Fang, Sam Buchanan, and Jeremias Sulam.
\newblock What's in a prior? learned proximal networks for inverse problems.
\newblock In \emph{The Twelfth International Conference on Learning
  Representations}, 2024.

\bibitem[Fang et~al.(2025)Fang, D{\'\i}az, Buchanan, and Sulam]{fang2025beyond}
Zhenghan Fang, Mateo D{\'\i}az, Sam Buchanan, and Jeremias Sulam.
\newblock Beyond scores: Proximal diffusion models.
\newblock \emph{arXiv preprint arXiv:2507.08956}, 2025.

\bibitem[Girrbach et~al.(2025)Girrbach, Alaniz, Smith, Darrell, and
  Akata]{girrbach2025person}
Leander Girrbach, Stephan Alaniz, Genevieve Smith, Trevor Darrell, and Zeynep
  Akata.
\newblock Person-centric annotations of laion-400m: Auditing bias and its
  transfer to models.
\newblock \emph{arXiv preprint arXiv:2510.03721}, 2025.

\bibitem[Ho and Salimans(2021)]{ho2021classifier}
Jonathan Ho and Tim Salimans.
\newblock Classifier-free diffusion guidance.
\newblock In \emph{NeurIPS 2021 Workshop on Deep Generative Models and
  Downstream Applications}, 2021.

\bibitem[Ho and Salimans(2022)]{ho2022classifier}
Jonathan Ho and Tim Salimans.
\newblock Classifier-free diffusion guidance, 2022.

\bibitem[Ho et~al.(2020)Ho, Jain, and Abbeel]{ho2020denoising}
Jonathan Ho, Ajay Jain, and Pieter Abbeel.
\newblock Denoising diffusion probabilistic models.
\newblock In \emph{Advances in Neural Information Processing Systems
  (NeurIPS)}, 2020.

\bibitem[Holderrieth et~al.(2025)Holderrieth, Singer, Jaakkola, Chen, Lipman,
  and Karrer]{holderrieth2025glass}
Peter Holderrieth, Uriel Singer, Tommi Jaakkola, Ricky T.~Q. Chen, Yaron
  Lipman, and Brian Karrer.
\newblock Glass flows: Transition sampling for alignment of flow and diffusion
  models, 2025.

\bibitem[Hu et~al.(2025)Hu, Zhang, Chen, Kuang, Li, Gao, Xiao, Wang, and
  Zhu]{hu2025towards}
Zijing Hu, Fengda Zhang, Long Chen, Kun Kuang, Jiahui Li, Kaifeng Gao, Jun
  Xiao, Xin Wang, and Wenwu Zhu.
\newblock Towards better alignment: Training diffusion models with
  reinforcement learning against sparse rewards.
\newblock In \emph{Proceedings of the Computer Vision and Pattern Recognition
  Conference}, pages 23604--23614, 2025.

\bibitem[Ju et~al.(2023)Ju, Zeng, Wang, Xu, and Zhang]{ju2023human}
Xuan Ju, Ailing Zeng, Jianan Wang, Qiang Xu, and Lei Zhang.
\newblock Human-art: A versatile human-centric dataset bridging natural and
  artificial scenes.
\newblock In \emph{Proceedings of the IEEE/CVF conference on computer vision
  and pattern recognition}, pages 618--629, 2023.

\bibitem[Karras et~al.(2022)Karras, Aittala, Aila, and
  Laine]{karras2022elucidating}
Tero Karras, Miika Aittala, Timo Aila, and Samuli Laine.
\newblock Elucidating the design space of diffusion-based generative models.
\newblock \emph{Advances in neural information processing systems},
  35:\penalty0 26565--26577, 2022.

\bibitem[Kingma and Welling(2014)]{kingma2014suto}
Diederik~P. Kingma and Max Welling.
\newblock {Auto-Encoding Variational Bayes}.
\newblock In \emph{2nd International Conference on Learning Representations,
  {ICLR} 2014, Banff, AB, Canada, April 14-16, 2014, Conference Track
  Proceedings}, 2014.

\bibitem[Kirstain et~al.(2023)Kirstain, Polyak, Singer, Matiana, Penna, and
  Levy]{kirstain2023pick}
Yuval Kirstain, Adam Polyak, Uriel Singer, Shahbuland Matiana, Joe Penna, and
  Omer Levy.
\newblock Pick-a-pic: An open dataset of user preferences for text-to-image
  generation.
\newblock \emph{Advances in neural information processing systems},
  36:\penalty0 36652--36663, 2023.

\bibitem[Li et~al.(2024)Li, Fu, Liu, Wang, Lin, and Wu]{li2024cosmicman}
Shikai Li, Jianglin Fu, Kaiyuan Liu, Wentao Wang, Kwan-Yee Lin, and Wayne Wu.
\newblock Cosmicman: A text-to-image foundation model for humans.
\newblock In \emph{Proceedings of the IEEE/CVF Conference on Computer Vision
  and Pattern Recognition}, pages 6955--6965, 2024.

\bibitem[Liao et~al.(2024)Liao, Xie, Chen, Lu, and Deng]{liao2024facescore}
Zhenyi Liao, Qingsong Xie, Chen Chen, Hannan Lu, and Zhijie Deng.
\newblock Facescore: Benchmarking and enhancing face quality in human
  generation.
\newblock \emph{arXiv preprint arXiv:2406.17100}, 2024.

\bibitem[Lipman et~al.(2022)Lipman, Chen, Ben-Hamu, Nickel, and
  Le]{lipman2022flow}
Yaron Lipman, Ricky~TQ Chen, Heli Ben-Hamu, Maximilian Nickel, and Matthew Le.
\newblock Flow matching for generative modeling.
\newblock \emph{arXiv preprint arXiv:2210.02747}, 2022.

\bibitem[Liu et~al.(2025)Liu, Liu, Liang, Li, Liu, Wang, Wan, Zhang, and
  Ouyang]{liu2025flow}
Jie Liu, Gongye Liu, Jiajun Liang, Yangguang Li, Jiaheng Liu, Xintao Wang,
  Pengfei Wan, Di Zhang, and Wanli Ouyang.
\newblock Flow-grpo: Training flow matching models via online rl.
\newblock \emph{arXiv preprint arXiv:2505.05470}, 2025.

\bibitem[Liu et~al.(2023)Liu, Gong, and qiang liu]{liu2023flow}
Xingchao Liu, Chengyue Gong, and qiang liu.
\newblock Flow straight and fast: Learning to generate and transfer data with
  rectified flow.
\newblock In \emph{The Eleventh International Conference on Learning
  Representations}, 2023.

\bibitem[Lu et~al.(2022)Lu, Zhou, Bao, Chen, Li, and Zhu]{lu2022dpm}
Cheng Lu, Yuhao Zhou, Fan Bao, Jianfei Chen, Chongxuan Li, and Jun Zhu.
\newblock Dpm-solver: A fast ode solver for diffusion probabilistic model
  sampling in around 10 steps.
\newblock \emph{Advances in neural information processing systems},
  35:\penalty0 5775--5787, 2022.

\bibitem[Lu et~al.(2025)Lu, Zhou, Bao, Chen, Li, and Zhu]{lu2025dpm}
Cheng Lu, Yuhao Zhou, Fan Bao, Jianfei Chen, Chongxuan Li, and Jun Zhu.
\newblock Dpm-solver++: Fast solver for guided sampling of diffusion
  probabilistic models.
\newblock \emph{Machine Intelligence Research}, pages 1--22, 2025.

\bibitem[Luhman and Luhman(2021)]{luhman2021knowledge}
Eric Luhman and Troy Luhman.
\newblock Knowledge distillation in iterative generative models for improved
  sampling speed.
\newblock \emph{arXiv preprint arXiv:2101.02388}, 2021.

\bibitem[Nichol et~al.(2021)Nichol, Dhariwal, Ramesh, Shyam, Mishkin, McGrew,
  Sutskever, and Chen]{nichol2021glide}
Alex Nichol, Prafulla Dhariwal, Aditya Ramesh, Pranav Shyam, Pamela Mishkin,
  Bob McGrew, Ilya Sutskever, and Mark Chen.
\newblock Glide: Towards photorealistic image generation and editing with
  text-guided diffusion models.
\newblock In \emph{International Conference on Machine Learning}, 2021.

\bibitem[Oksendal(2013)]{oksendal2013stochastic}
Bernt Oksendal.
\newblock \emph{Stochastic differential equations: an introduction with
  applications}.
\newblock Springer Science \& Business Media, 2013.

\bibitem[Podell et~al.(2024)Podell, English, Lacey, Blattmann, Dockhorn,
  M{\"u}ller, Penna, and Rombach]{podell2024sdxl}
Dustin Podell, Zion English, Kyle Lacey, Andreas Blattmann, Tim Dockhorn, Jonas
  M{\"u}ller, Joe Penna, and Robin Rombach.
\newblock {SDXL}: Improving latent diffusion models for high-resolution image
  synthesis.
\newblock In \emph{The Twelfth International Conference on Learning
  Representations}, 2024.

\bibitem[Qwen et~al.(2025)Qwen, :, Yang, Yang, Zhang, Hui, Zheng, Yu, Li, Liu,
  Huang, Wei, Lin, Yang, Tu, Zhang, Yang, Yang, Zhou, Lin, Dang, Lu, Bao, Yang,
  Yu, Li, Xue, Zhang, Zhu, Men, Lin, Li, Tang, Xia, Ren, Ren, Fan, Su, Zhang,
  Wan, Liu, Cui, Zhang, and Qiu]{qwen2025qwen25technicalreport}
Qwen, :, An Yang, Baosong Yang, Beichen Zhang, Binyuan Hui, Bo Zheng, Bowen Yu,
  Chengyuan Li, Dayiheng Liu, Fei Huang, Haoran Wei, Huan Lin, Jian Yang,
  Jianhong Tu, Jianwei Zhang, Jianxin Yang, Jiaxi Yang, Jingren Zhou, Junyang
  Lin, Kai Dang, Keming Lu, Keqin Bao, Kexin Yang, Le Yu, Mei Li, Mingfeng Xue,
  Pei Zhang, Qin Zhu, Rui Men, Runji Lin, Tianhao Li, Tianyi Tang, Tingyu Xia,
  Xingzhang Ren, Xuancheng Ren, Yang Fan, Yang Su, Yichang Zhang, Yu Wan,
  Yuqiong Liu, Zeyu Cui, Zhenru Zhang, and Zihan Qiu.
\newblock Qwen2.5 technical report, 2025.

\bibitem[Raffel et~al.(2020)Raffel, Shazeer, Roberts, Lee, Narang, Matena,
  Zhou, Li, and Liu]{raffel2020exploring}
Colin Raffel, Noam Shazeer, Adam Roberts, Katherine Lee, Sharan Narang, Michael
  Matena, Yanqi Zhou, Wei Li, and Peter~J Liu.
\newblock Exploring the limits of transfer learning with a unified text-to-text
  transformer.
\newblock \emph{Journal of machine learning research}, 21\penalty0
  (140):\penalty0 1--67, 2020.

\bibitem[Ramesh et~al.(2022)Ramesh, Pavlov, Goh, Gray, Voss, Radford, Chen, and
  Sutskever]{ramesh2022hierarchical}
Aditya Ramesh, Mikhail Pavlov, Gabriel Goh, Scott Gray, Chelsea Voss, Alec
  Radford, Mark Chen, and Ilya Sutskever.
\newblock Hierarchical text-conditional image generation with clip latents.
\newblock \emph{arXiv preprint arXiv:2204.06125}, 2022.

\bibitem[Rombach et~al.(2022)Rombach, Blattmann, Lorenz, Esser, and
  Ommer]{rombach2022ldm}
Robin Rombach, Andreas Blattmann, Dominik Lorenz, Patrick Esser, and Bj{\"o}rn
  Ommer.
\newblock High-resolution image synthesis with latent diffusion models.
\newblock In \emph{IEEE Conference on Computer Vision and Pattern Recognition
  (CVPR)}, 2022.

\bibitem[Saharia et~al.(2022)Saharia, Chan, Saxena, Li, Whang, Denton,
  Ghasemipour, Lopes, and et~al.]{saharia2022imagen}
Chitwan Saharia, William Chan, Saurabh Saxena, Lala Li, Jay Whang, Emily
  Denton, Seyed~Kamyar Ghasemipour, Raphael Lopes, and et al.
\newblock Photorealistic text-to-image diffusion models with deep language
  understanding.
\newblock In \emph{Advances in Neural Information Processing Systems
  (NeurIPS)}, 2022.

\bibitem[Salimans and Ho(2022)]{salimans2022progressive}
Tim Salimans and Jonathan Ho.
\newblock Progressive distillation for fast sampling of diffusion models.
\newblock In \emph{International Conference on Learning Representations
  (ICLR)}, 2022.

\bibitem[Schuhmann(2022)]{schuhmann2022laionaesthetics}
Christoph Schuhmann.
\newblock {LAION-Aesthetics}.
\newblock \url{https://laion.ai/blog/laion-aesthetics/}, 2022.
\newblock Accessed: 2025-11-08.

\bibitem[Schulman et~al.(2015)Schulman, Levine, Abbeel, Jordan, and
  Moritz]{schulman2015trust}
John Schulman, Sergey Levine, Pieter Abbeel, Michael Jordan, and Philipp
  Moritz.
\newblock Trust region policy optimization.
\newblock In \emph{Proceedings of the 32nd International Conference on Machine
  Learning}, pages 1889--1897, Lille, France, 2015. PMLR.

\bibitem[Schulman et~al.(2017)Schulman, Wolski, Dhariwal, Radford, and
  Klimov]{schulman2017proximal}
John Schulman, Filip Wolski, Prafulla Dhariwal, Alec Radford, and Oleg Klimov.
\newblock Proximal policy optimization algorithms, 2017.

\bibitem[Shao et~al.(2024)Shao, Wang, Zhu, Xu, Song, Bi, Zhang, Zhang, Li, Wu,
  et~al.]{shao2024deepseekmath}
Zhihong Shao, Peiyi Wang, Qihao Zhu, Runxin Xu, Junxiao Song, Xiao Bi, Haowei
  Zhang, Mingchuan Zhang, YK Li, Yang Wu, et~al.
\newblock Deepseekmath: Pushing the limits of mathematical reasoning in open
  language models.
\newblock \emph{arXiv preprint arXiv:2402.03300}, 2024.

\bibitem[Sohl-Dickstein et~al.(2015)Sohl-Dickstein, Weiss, Maheswaranathan, and
  Ganguli]{sohl2015deep}
Jascha Sohl-Dickstein, Eric Weiss, Niru Maheswaranathan, and Surya Ganguli.
\newblock Deep unsupervised learning using nonequilibrium thermodynamics.
\newblock In \emph{International conference on machine learning}, pages
  2256--2265. pmlr, 2015.

\bibitem[Song et~al.(2021{\natexlab{a}})Song, Meng, and
  Ermon]{song2021denoising}
Jiaming Song, Chenlin Meng, and Stefano Ermon.
\newblock Denoising diffusion implicit models.
\newblock In \emph{International Conference on Learning Representations},
  2021{\natexlab{a}}.

\bibitem[Song and Ermon(2019)]{song2019generative}
Yang Song and Stefano Ermon.
\newblock Generative modeling by estimating gradients of the data distribution.
\newblock \emph{Advances in neural information processing systems}, 32, 2019.

\bibitem[Song et~al.(2021{\natexlab{b}})Song, Sohl-Dickstein, Kingma, Kumar,
  Ermon, and Poole]{song2021score}
Yang Song, Jascha Sohl-Dickstein, Diederik~P. Kingma, Abhishek Kumar, Stefano
  Ermon, and Ben Poole.
\newblock Score-based generative modeling through stochastic differential
  equations.
\newblock In \emph{International Conference on Learning Representations
  (ICLR)}, 2021{\natexlab{b}}.

\bibitem[Uehara et~al.(2024{\natexlab{a}})Uehara, Zhao, Biancalani, and
  Levine]{uehara2024understanding}
Masatoshi Uehara, Yulai Zhao, Tommaso Biancalani, and Sergey Levine.
\newblock Understanding reinforcement learning-based fine-tuning of diffusion
  models: A tutorial and review.
\newblock \emph{arXiv preprint arXiv:2407.13734}, 2024{\natexlab{a}}.

\bibitem[Uehara et~al.(2024{\natexlab{b}})Uehara, Zhao, Black, Hajiramezanali,
  Scalia, Diamant, Tseng, Biancalani, and Levine]{uehara2024fine}
Masatoshi Uehara, Yulai Zhao, Kevin Black, Ehsan Hajiramezanali, Gabriele
  Scalia, Nathaniel~Lee Diamant, Alex~M Tseng, Tommaso Biancalani, and Sergey
  Levine.
\newblock Fine-tuning of continuous-time diffusion models as
  entropy-regularized control.
\newblock \emph{arXiv preprint arXiv:2402.15194}, 2024{\natexlab{b}}.

\bibitem[Wu et~al.(2023)Wu, Hao, Sun, Chen, Zhu, Zhao, and Li]{wu2023human}
Xiaoshi Wu, Yiming Hao, Keqiang Sun, Yixiong Chen, Feng Zhu, Rui Zhao, and
  Hongsheng Li.
\newblock Human preference score v2: A solid benchmark for evaluating human
  preferences of text-to-image synthesis.
\newblock \emph{arXiv preprint arXiv:2306.09341}, 2023.

\bibitem[Wu et~al.(2024)Wu, Chen, and Wei]{wu2024stochastic}
Yuchen Wu, Yuxin Chen, and Yuting Wei.
\newblock Stochastic runge-kutta methods: Provable acceleration of diffusion
  models.
\newblock \emph{arXiv preprint arXiv:2410.04760}, 2024.

\bibitem[Xia et~al.(2021{\natexlab{a}})Xia, Yang, Xue, and Wu]{xia2021tedigan}
Weihao Xia, Yujiu Yang, Jing-Hao Xue, and Baoyuan Wu.
\newblock Tedigan: Text-guided diverse face image generation and manipulation.
\newblock In \emph{IEEE Conference on Computer Vision and Pattern Recognition
  (CVPR)}, 2021{\natexlab{a}}.

\bibitem[Xia et~al.(2021{\natexlab{b}})Xia, Yang, Xue, and Wu]{xia2021towards}
Weihao Xia, Yujiu Yang, Jing-Hao Xue, and Baoyuan Wu.
\newblock Towards open-world text-guided face image generation and
  manipulation.
\newblock \emph{arxiv preprint arxiv: 2104.08910}, 2021{\natexlab{b}}.

\bibitem[Xu et~al.(2023{\natexlab{a}})Xu, Xie, Tan, Huang, Howes, Sharma, Li,
  Ghosh, Zettlemoyer, and Feichtenhofer]{xu2023demystifying}
Hu Xu, Saining Xie, Xiaoqing~Ellen Tan, Po-Yao Huang, Russell Howes, Vasu
  Sharma, Shang-Wen Li, Gargi Ghosh, Luke Zettlemoyer, and Christoph
  Feichtenhofer.
\newblock Demystifying clip data.
\newblock \emph{arXiv preprint arXiv:2309.16671}, 2023{\natexlab{a}}.

\bibitem[Xu et~al.(2023{\natexlab{b}})Xu, Liu, Wu, Tong, Li, Ding, Tang, and
  Dong]{xu2023imagereward}
Jiazheng Xu, Xiao Liu, Yuchen Wu, Yuxuan Tong, Qinkai Li, Ming Ding, Jie Tang,
  and Yuxiao Dong.
\newblock Imagereward: Learning and evaluating human preferences for
  text-to-image generation.
\newblock \emph{Advances in Neural Information Processing Systems},
  36:\penalty0 15903--15935, 2023{\natexlab{b}}.

\bibitem[Xue et~al.(2025)Xue, Wu, Gao, Kong, Zhu, Chen, Liu, Liu, Guo, Huang,
  et~al.]{xue2025dancegrpo}
Zeyue Xue, Jie Wu, Yu Gao, Fangyuan Kong, Lingting Zhu, Mengzhao Chen, Zhiheng
  Liu, Wei Liu, Qiushan Guo, Weilin Huang, et~al.
\newblock Dancegrpo: Unleashing grpo on visual generation.
\newblock \emph{arXiv preprint arXiv:2505.07818}, 2025.

\bibitem[Zhang and Chen(2023)]{zhang2023fast}
Qinsheng Zhang and Yongxin Chen.
\newblock Fast sampling of diffusion models with exponential integrator.
\newblock In \emph{The Eleventh International Conference on Learning
  Representations}, 2023.

\end{thebibliography}
}

\newpage
\appendix
\onecolumn

\section{Full definition of the GRPO objective}
\label{sec:full-grpo-objective}
The full GRPO objective used in our work is defined as:
\begin{align*}
&\mathcal{J}_{\mathrm{GRPO}}(\theta) = 
 \sum_{i=1}^G \sum_{k=1}^K 
\left\{ 
R_k^{(i)} (\theta) - \beta_{\mathrm{KL}} D_{\mathrm{KL},k}^{(i)}(\theta)
\right\} ,
\end{align*}
where
\begin{align*}
\label{eq:clipped-reward}
R_k^{(i)} (\theta) = \min \left[ r_k^{(i)}(\theta) \hat{A}^{(i)} , \operatorname{clip} \left( r_k^{(i)}(\theta) , 1 - \epsilon, 1+ \epsilon \right) \hat{A}^{(i)} \right],
\end{align*}
and
\begin{align*}
    r^i_k(\theta) = \frac{p_\theta (X^{(i)}_{k-1} \mid X^{(i)}_{k}, c)}{p_{\theta_{\text{old}}}(X^{(i)}_{k-1} \mid X^{(i)}_{k}, c)}.
\end{align*}
This clipped reward imposes a trust-region–like constraint to avoid large policy updates and stabilizes training under noisy reward estimates \citep{schulman2015trust,schulman2017proximal}.

\section{Processing Details of the LAION-Face-T2I-15M Dataset}
\label{sec:appx-t2i-dataset}

We start from the LAION-Face dataset\footnote{\url{https://github.com/FacePerceiver/LAION-Face}} and apply additional filtering to retain images with a minimum short edge of $\ge256$ pixels, aspect ratio in $[9/16,16/9]$, and an aesthetic score $\ge4.5$ (the LAION aesthetic predictor \citep{schuhmann2022laionaesthetics}), resulting in a high-resolution and high-quality collection of roughly 15M images.

Since the raw LAION-Face captions are limited in details and not suitable for fine-grained text-to-image training, we recaption all images using the Qwen-2.5-7B vision-language model \citep{qwen2025qwen25technicalreport} with the following prompt:
\if\isArxiv 1
\lstset{basicstyle=\ttfamily\small,breaklines=true,columns=fullflexible,breakindent=0em}
\begin{lstlisting}
Write a concise summary of the image in 30 words. Describe the main subjects, actions, and any noticeable details such as face, hair, clothes, ethnicity, or background. Describe the image style, such as photography, painting, or cartoon. The description should be a suitable image generation prompt.
\end{lstlisting}
\fi
\if\isArxiv 0
\begin{minted}[fontsize=\small,breaklines=true]{text}
Write a concise summary of the image in 30 words. Describe the main subjects, actions, and any noticeable details such as face, hair, clothes, ethnicity, or background. Describe the image style, such as photography, painting, or cartoon. The description should be a suitable image generation prompt.
\end{minted}
\fi
This process yields high-quality captions suitable for training human text-to-image models from scratch.

In addition to the base dataset, we also construct LAION-Face-Hand-3M, a subset of LAION-Face-T2I-15M tailored for hand image generation. We employ Qwen-2.5-7B to identify images containing close-up views of hands and generate detailed captions describing the hand attributes, with the prompt shown below:
\if\isArxiv 1
\begin{lstlisting}
Does this image meet **all** of the following criteria?
1. It is a **real photograph**, not a drawing or digital rendering.
2. It shows a **close-up view** of the upper body (head to waist).
3. There is a **single person** in the image.
4. At least **one hand** is clearly visible and engaged in a motion or gesture (e.g., pointing, holding, touching, etc.).

If yes, respond with 'Yes' and on a new line, a caption of the image in about 30 words. Describe the hands in detail, covering:
- Number of hands visible (e.g., 'one hand', 'both hands')
- Position relative to the body (e.g., 'resting on lap', 'raised near face')
- Finger gesture or configuration (e.g., 'fingers curled', 'spread apart', 'pointing')
- Activity or interaction if any (e.g., 'adjusting clothing', 'touching chin', 'gesturing toward camera')
- Visibility quality (e.g., 'clearly visible', 'partially obscured').

If no, respond with 'No' and nothing else.
\end{lstlisting}
\fi
\if\isArxiv 0
\begin{minted}[fontsize=\small,breaklines=true]{text}
Does this image meet **all** of the following criteria?
1. It is a **real photograph**, not a drawing or digital rendering.
2. It shows a **close-up view** of the upper body (head to waist).
3. There is a **single person** in the image.
4. At least **one hand** is clearly visible and engaged in a motion or gesture (e.g., pointing, holding, touching, etc.).

If yes, respond with 'Yes' and on a new line, a caption of the image in about 30 words. Describe the hands in detail, covering:
- Number of hands visible (e.g., 'one hand', 'both hands')
- Position relative to the body (e.g., 'resting on lap', 'raised near face')
- Finger gesture or configuration (e.g., 'fingers curled', 'spread apart', 'pointing')
- Activity or interaction if any (e.g., 'adjusting clothing', 'touching chin', 'gesturing toward camera')
- Visibility quality (e.g., 'clearly visible', 'partially obscured').

If no, respond with 'No' and nothing else.
\end{minted}
\fi
The resulting captions describe the number of hands, their position relative to the body, finger gestures, interactions (e.g., holding objects or touching the face), and visibility quality. This level of annotation provides training data with a much finer level of supervision for learning human-hand generation, an area where current models often fail.

\begin{figure}[h]
\centering
\includegraphics[width=\linewidth]{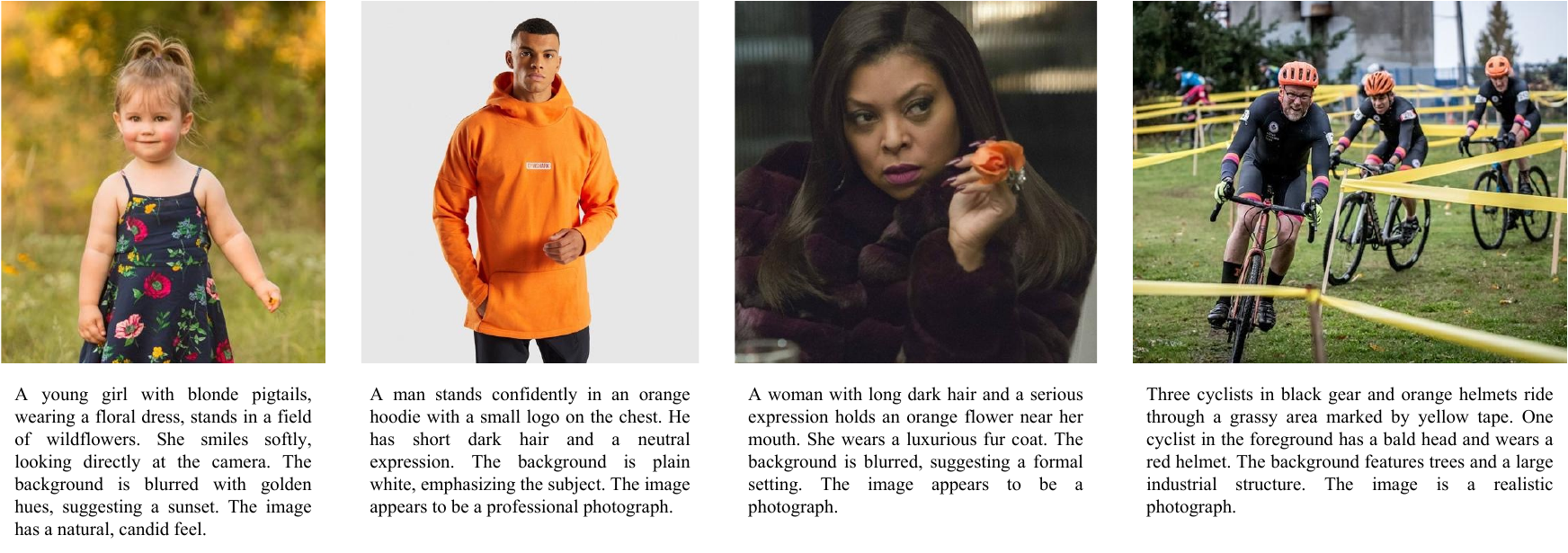}
\caption{Example images and corresponding fine-grained captions in the LAION-Face-T2I-15M dataset.}
\label{fig:dataset-examples}
\end{figure}

\section{Experimental Details}
\label{sec:appx-experimental-details}

\paragraph{Implementation.}
To enable efficient sampling of high-resolution images, we perform generation in the latent space of a pretrained variational autoencoder (VAE) \citep{kingma2014suto,rombach2022ldm}, which compresses images into a lower-dimensional representation. Specifically, we use the VAE from Stable Diffusion 3.5 \citep{esser2024scaling}, which encodes an image of size $H \times W \times 3$ to a latent dimension of $H/8 \times W/8 \times 16$.

For textual conditioning, we leverage a combination of pretrained text encoders to extract semantically rich and diverse representations of the input prompt. In particular, we use the MetaCLIP encoder \citep{xu2023demystifying} trained on openly curated data %
and the T5 encoder \citep{raffel2020exploring}, which has been observed to benefit generation for complex prompts involving dense details \citep{esser2024scaling}, aligning well with our objective of fine-grained human image generation. The outputs from the two encoders are combined by padding to a common embedding dimension and then concatenating along the token axis. 

For the proximal network $f_\theta(x; t, \lambda, c)$, we adopt the U-ViT architecture \citep{bao2023all}--a Vision Transformer (ViT) \citep{dosovitskiy2021an} based backbone--for its flexibility in handling variation in the input space by treating all inputs as tokens. This network design naturally accommodates the additional $\lambda$-conditioning present in the proximal model but absent in conventional score networks. The joint distribution of $(t, \lambda)$ during training is defined according to the proximal operators required for a range of potential sampling steps ($\{4,5,6,7,8,9,10,25\}$), to target efficient sampling.
This conditional proximal diffusion model serves as the foundation for the reinforcement learning stage described next.

\paragraph{Training.} Our models are trained from scratch on $256^2$ images and their corresponding captions from \textsc{LAION-Face-T2I-15M}. We train a score-based and a proximal diffusion model, both using a batch size of 2,048 and for 100 thousand steps. We train ProxT2I using proximal matching loss, with a learning rate of $8\times10^{-5}$. We set $\zeta=1.0$ in the proximal matching loss, and sample $(t,\lambda)$ according to the discretization schemes corresponding to the sampling steps: $\{4, 5, 6, 7, 8, 9, 10, 25\}$. The score models are trained using the score matching loss \citep{song2019generative,ho2020denoising} with a learning rate of $4\times10^{-4}$ following \citep{bao2023all}. We randomly set the input text conditioning to an empty string with a probability $p_{\varnothing}=0.1$ to train unconditional models for classifier-free guidance for both score and proximal models.

We then optimize the pretrained models for specific rewards using reinforcement learning with GRPO. We use PickScore \citep{kirstain2023pick} as the reward model and a training dataset contains 3,000 prompts. We draw samples using 10 sampling steps. We set the weight for the KL regularization to $\beta=0.001$. We set the group size to $G=24$ and the number of sampling steps to $K=10$. Each training batch contains 12 prompts. We use a gradient accumulation of 6 batches, and train each model for 1,000 steps.

To study the performance of ProxT2I for higher-resolution generation, we finetune the pretrained $256^2$ models on $512^2$ images from \textsc{LAION-Face-T2I-15M} and \textsc{LAION-Face-T2I-Hand-3M}. We also include images from MMCelebA-HQ \citep{xia2021tedigan} and CosmicMan \citep{li2024cosmicman} to improve training data diversity, with captions generated following the same procedures as \textsc{LAION-Face-T2I-15M} for all images. Images are filtered by short edge $\geq 512$ and aspect ratio in $[0.9,\,1/0.9]$. This stage uses a batch size of 256 and trains both score and ProxT2I for 60k steps, with a learning rate of $1\times10^{-5}$. Finally, both the score and proximal models are fine-tuned using GRPO on 100 curated prompts with PickScore as the reward model for 1,000 steps, with other settings the same as the $256^2$ experiment.

\section{Additional Results}

\begin{figure*}[h]
\centering
\begin{subfigure}[t]{0.49\linewidth}
    \centering
    \includegraphics[width=\linewidth]{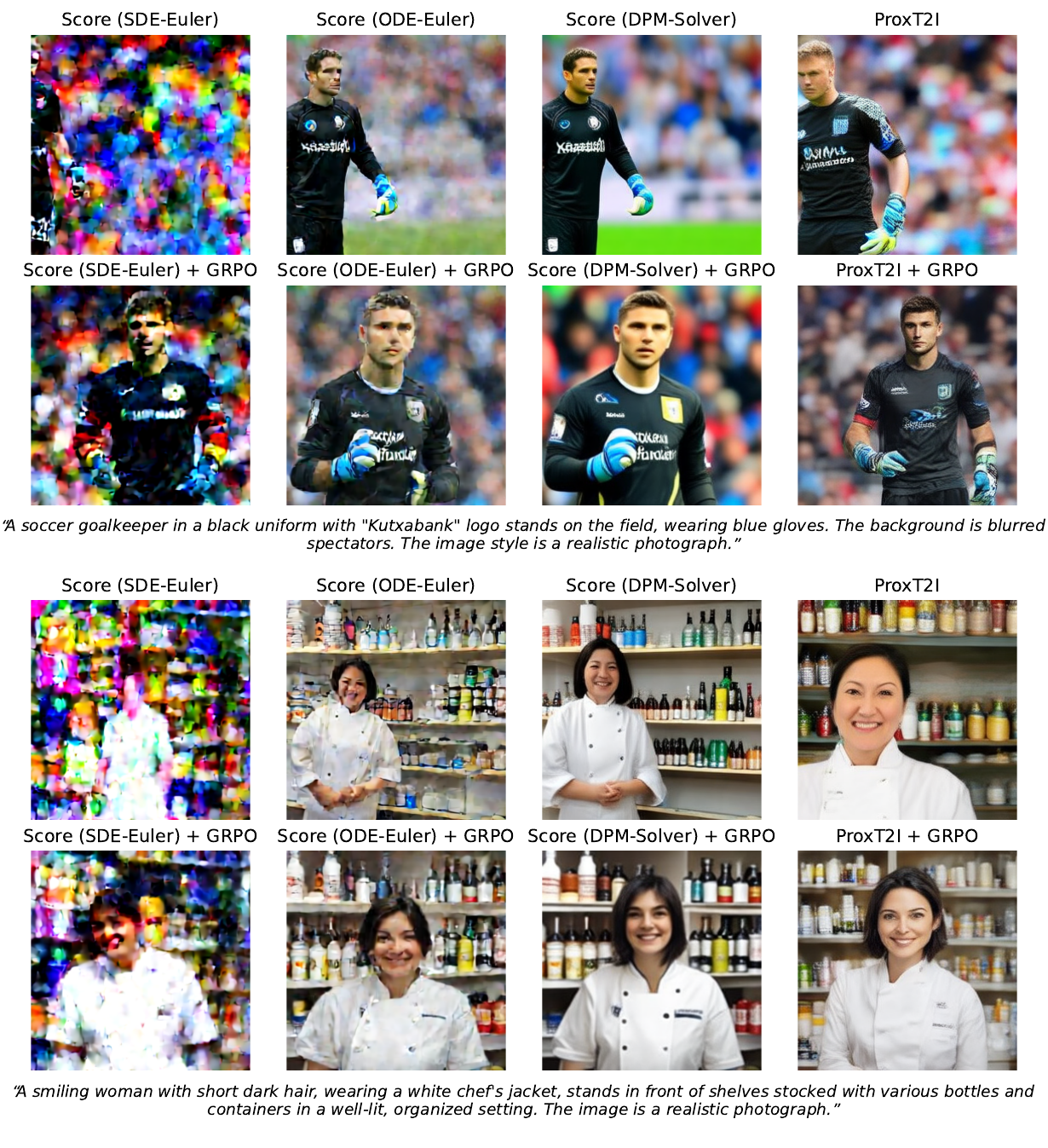}
    \caption{6 sampling steps.}
\end{subfigure}
\hfill
\begin{subfigure}[t]{0.49\linewidth}
    \centering
    \includegraphics[width=\linewidth]{figures/comparison_grid_256_10steps.pdf}
    \caption{10 sampling steps.}
\end{subfigure}

\caption{Qualitative comparison of generated images at $256^2$ resolution with 6 and 10 sampling steps.}
\label{fig:256-images}
\end{figure*}

\begin{figure*}[h]
    \centering
    \subcaptionbox{ImageReward}{%
        \includegraphics[width=0.45\textwidth]{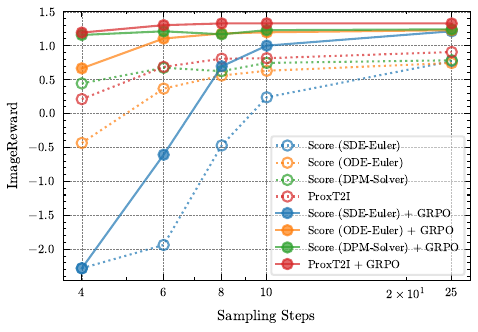}}
    \subcaptionbox{PickScore}{%
        \includegraphics[width=0.45\textwidth]{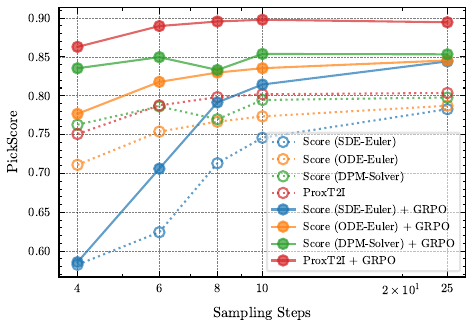}}
    \subcaptionbox{Aesthetic Score}{%
        \includegraphics[width=0.45\textwidth]{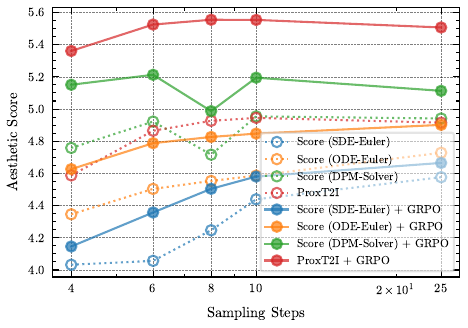}}
    \caption{Comparison of additional metrics at 256$^2$ resolution.}
    \label{fig:256-metric-appx}
\end{figure*}

\begin{figure}[t]
\centering
\includegraphics[width=.95\linewidth]{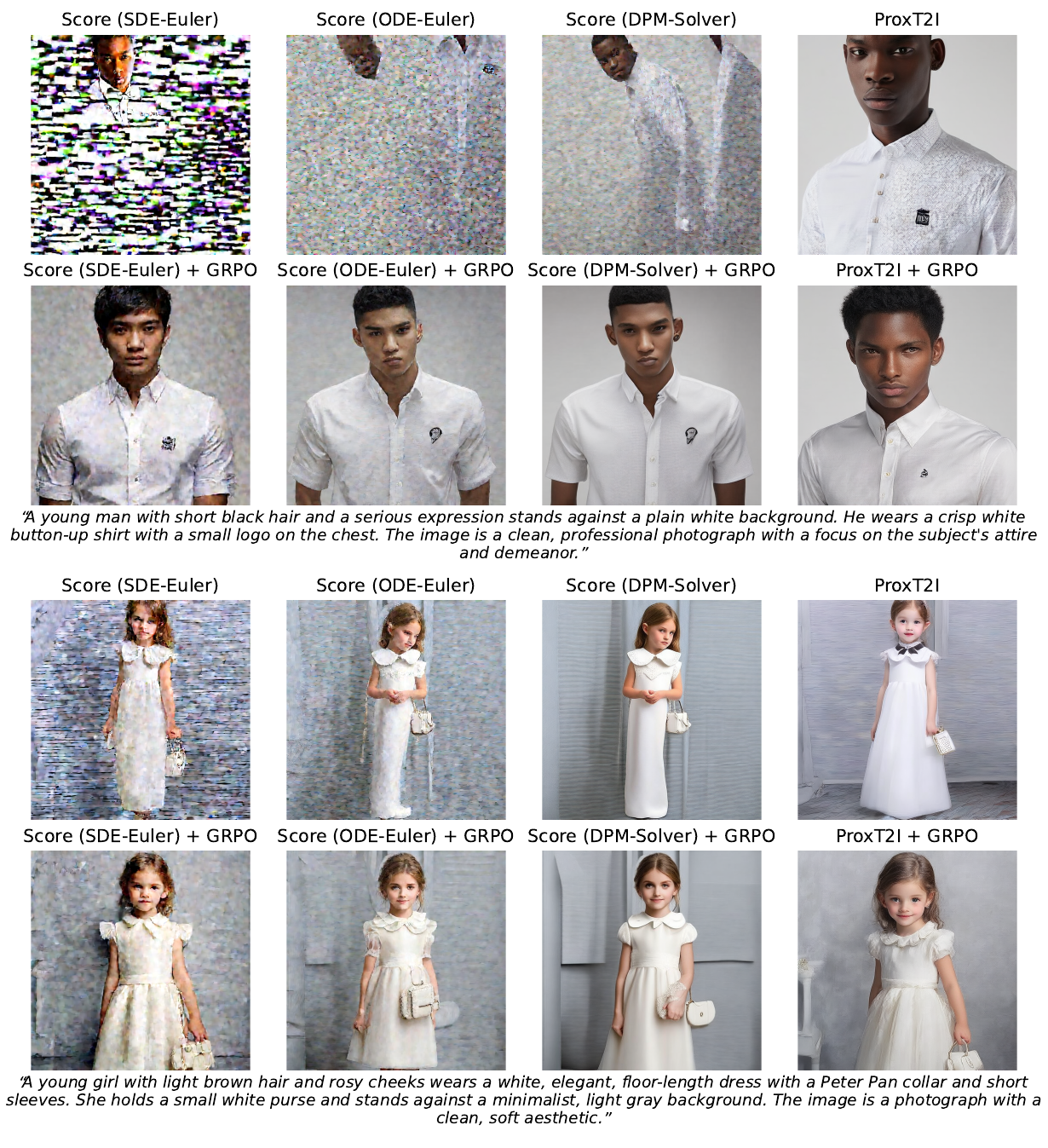}
\caption{Qualitative comparison of 512$^2$ generations at 10 sampling steps.}
\label{fig:512-images}
\end{figure}

\begin{figure*}[h]
    \centering
    \subcaptionbox{ImageReward}{%
        \includegraphics[width=0.45\textwidth]{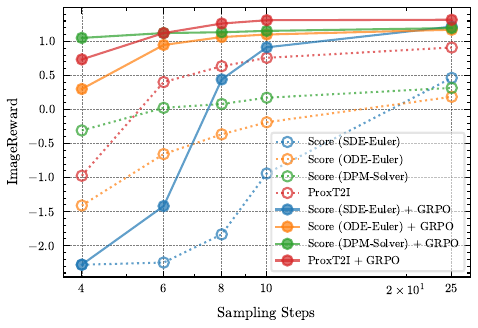}}
    \subcaptionbox{PickScore}{%
        \includegraphics[width=0.45\textwidth]{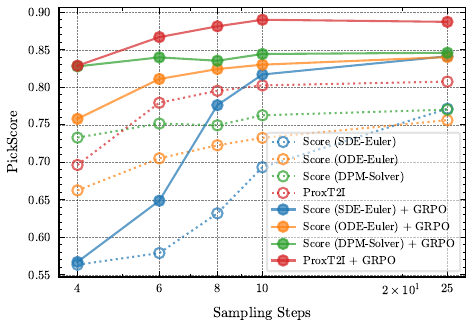}}
    \subcaptionbox{Aesthetic Score}{%
        \includegraphics[width=0.45\textwidth]{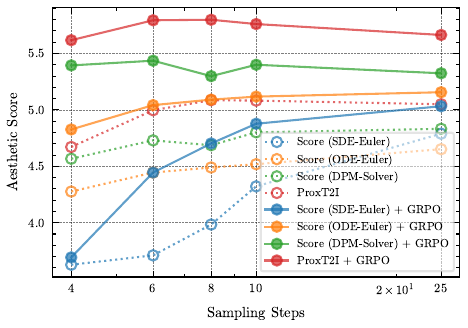}}
    \caption{Comparison of additional metrics  at $512^2$ resolution.}
    \label{fig:512-metric-appx}
\end{figure*}

\begin{figure*}[h]
    \centering
    \subcaptionbox{HPSv2.1, $256^2$}{%
        \includegraphics[width=0.4\textwidth]{figures/test_curve_external_prompts_1024_hpsv2.1.pdf}}
    \subcaptionbox{HPSv2.1, $512^2$}{%
        \includegraphics[width=0.4\textwidth]{figures/test_curve_512_external_prompts_1024_hpsv2.1.pdf}}
        
    \subcaptionbox{Aesthetic Score, $256^2$}{%
        \includegraphics[width=0.4\textwidth]{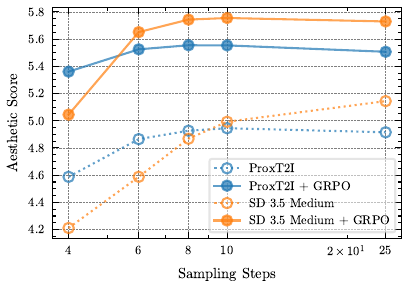}}
    \subcaptionbox{Aesthetic Score, $512^2$}{%
        \includegraphics[width=0.4\textwidth]{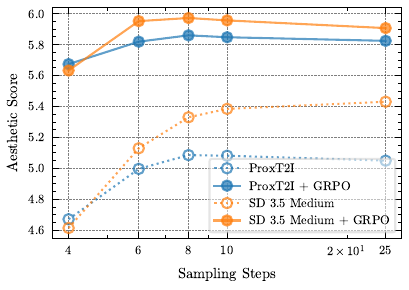}}
        
    \subcaptionbox{ImageReward, $256^2$}{%
        \includegraphics[width=0.4\textwidth]{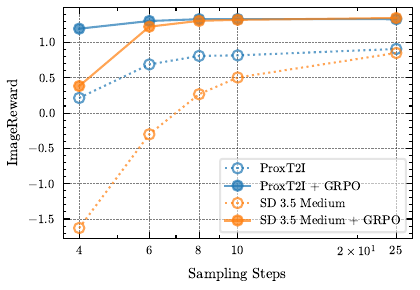}}
    \subcaptionbox{ImageReward, $512^2$}{%
        \includegraphics[width=0.4\textwidth]{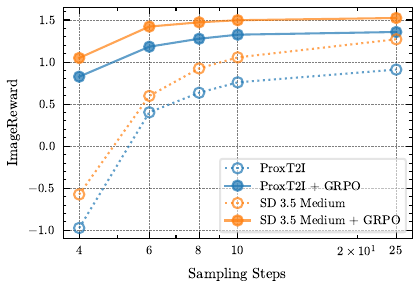}}
        
    \subcaptionbox{PickScore, $256^2$}{%
        \includegraphics[width=0.4\textwidth]{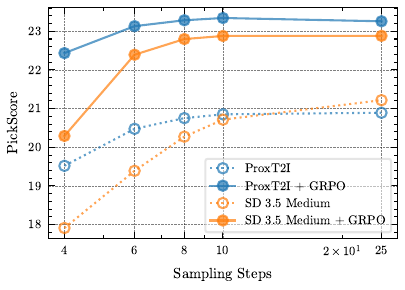}}
    \subcaptionbox{PickScore, $512^2$}{%
        \includegraphics[width=0.4\textwidth]{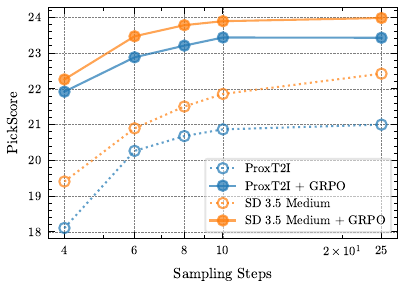}}
        
    \caption{Comparison with StableDiffusion (SD) 3.5 Medium \citep{esser2024scaling} and its RL finetuned model from FlowGRPO \citep{liu2025flow} at $256^2$ and $512^2$ resolutions.}
    \label{fig:external-appx}
\end{figure*}

\begin{figure*}[h]
    \centering
    \subcaptionbox{HPSv2.1}{
        \includegraphics[width=0.43\textwidth]{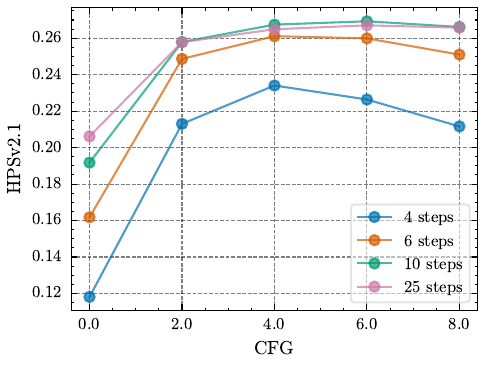}}
    \subcaptionbox{Aesthetic Score}{
        \includegraphics[width=0.43\textwidth]{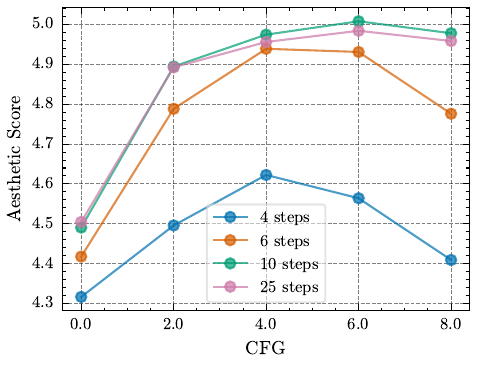}}
    \subcaptionbox{ImageReward}{
        \includegraphics[width=0.43\textwidth]{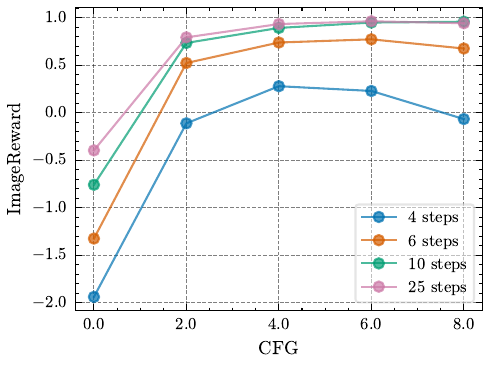}}
    \subcaptionbox{PickScore}{
        \includegraphics[width=0.43\textwidth]{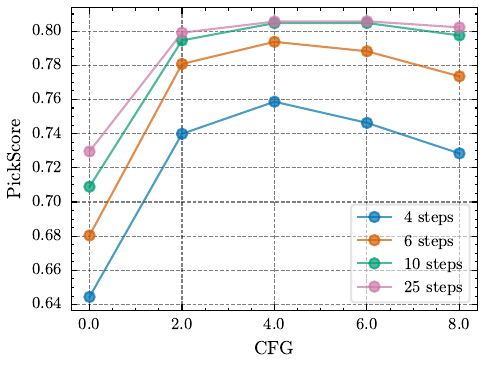}}
    \caption{Effect of CFG level $\omega$ for ProxT2I at different sampling step numbers. Here, CFG=0 corresponds to using no guidance, i.e., using only the conditional proximal operator without the unconditional one.}
    \label{fig:cfg}
\end{figure*}

\begin{figure}[t]
\centering
\includegraphics[width=.9\linewidth]{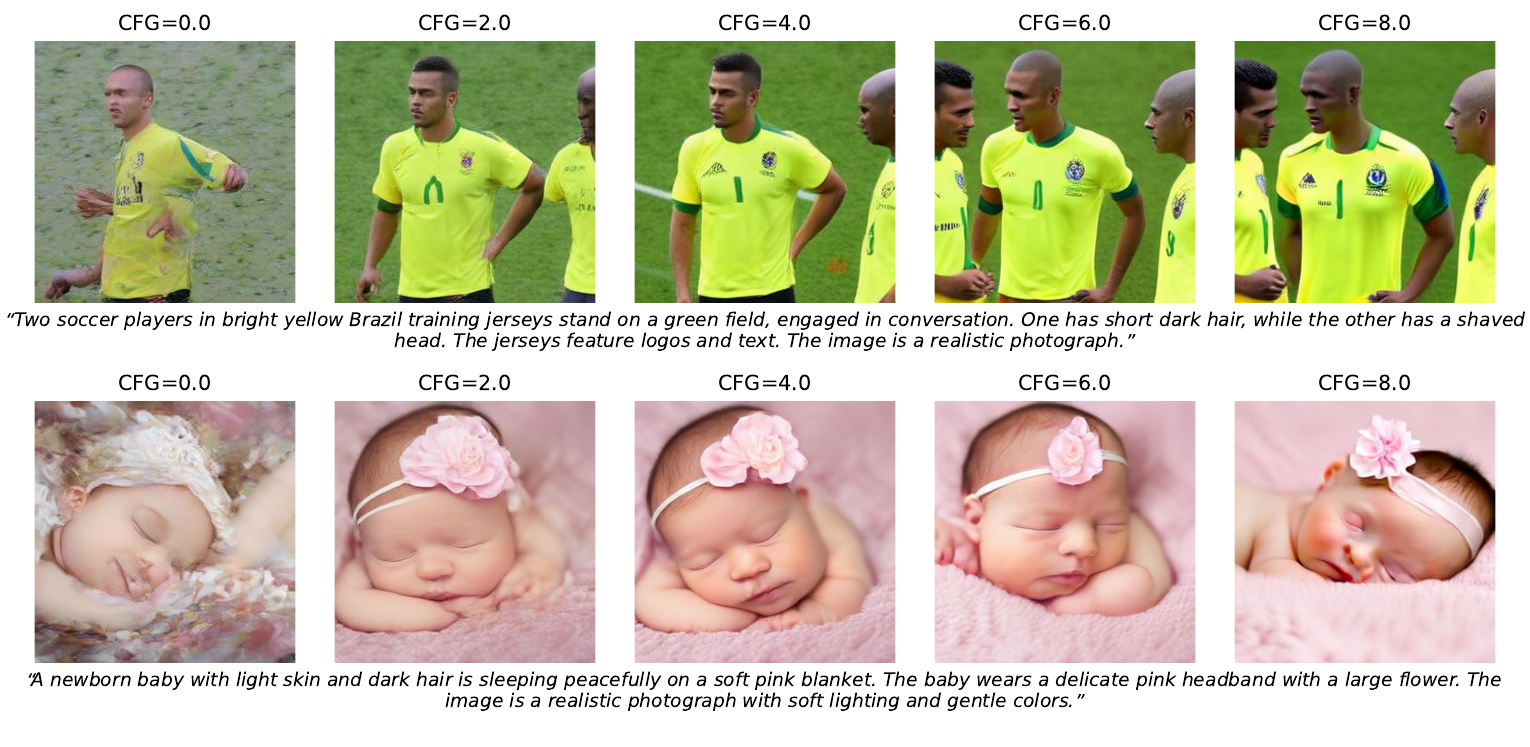}
\caption{Effect of CFG level $\omega$ on ProxT2I-generated images ($256^2$ resolution and 10 sampling steps). Increasing CFG level from 0 improves image quality and text-image alignment, while overly large values yield degraded result.}
\label{fig:cfg-images}
\end{figure}

\begin{figure*}[h]
    \centering
    \subcaptionbox{PickScore}{
        \includegraphics[width=0.24\textwidth]{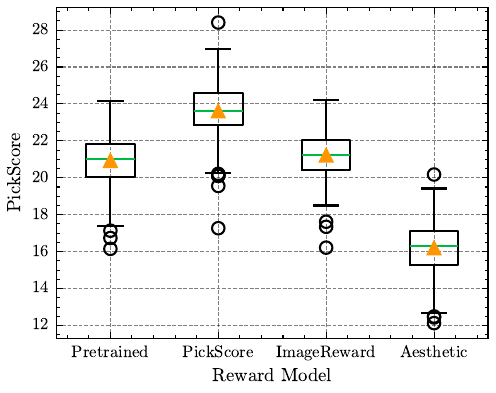}}
    \subcaptionbox{ImageReward}{
        \includegraphics[width=0.24\textwidth]{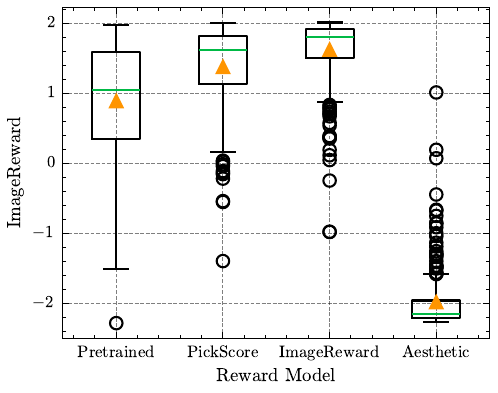}}
    \subcaptionbox{Aesthetic Score}{
        \includegraphics[width=0.24\textwidth]{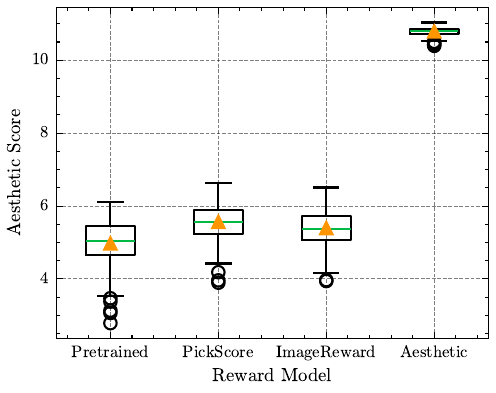}}
    \subcaptionbox{HPSv2.1}{
        \includegraphics[width=0.24\textwidth]{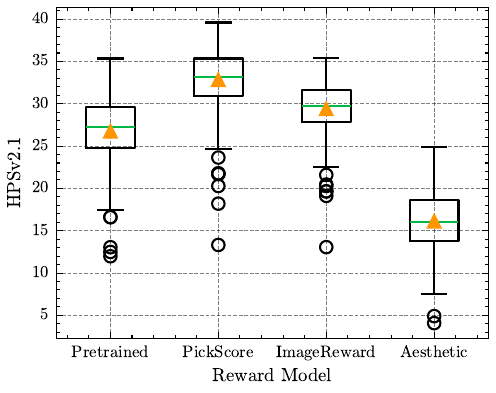}}
    \caption{Comparison of ProxT2I models finetuned with different reward functions. We apply reinforcement learning (RL) using three reward models: PickScore, ImageReward, and Aesthetic Score, and compare them with the pretrained (non-RL) baseline. Each panel shows the results of all models for a specific evaluation metric. Besides the three training rewards, we include HPSv2.1, an external metric not used in RL, to assess generalization. Results are summarized over 256 testing prompts and the yellow triangle denotes the mean. All results are obtained with 10 sampling steps at $256^2$ resolution.}
    \label{fig:ablation-reward}
\end{figure*}

\begin{figure*}
    \centering
    \includegraphics[width=0.8\textwidth]{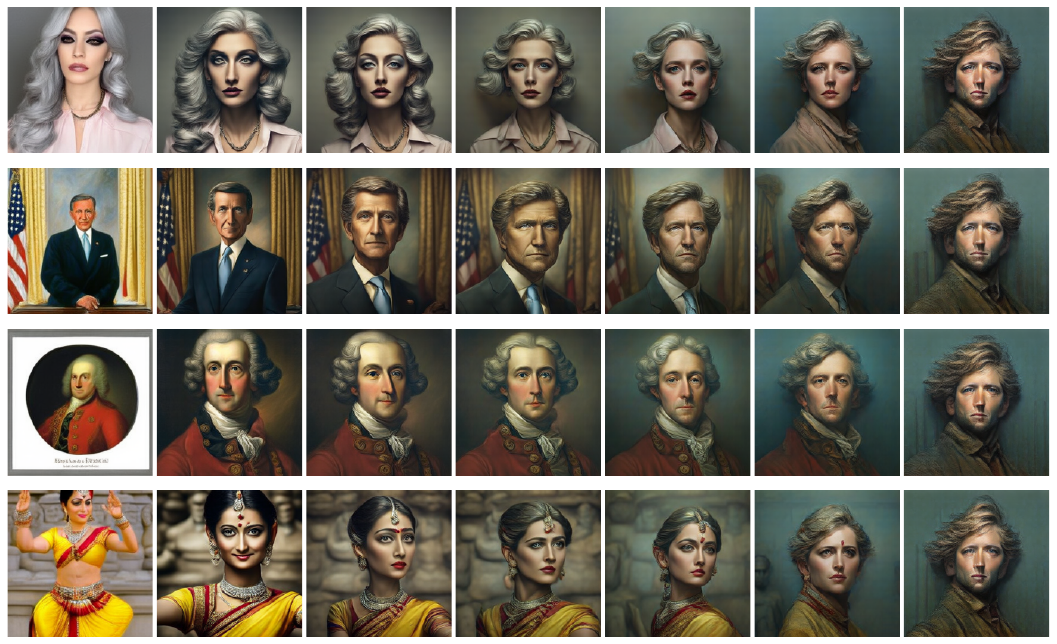}
    \caption{Visualization of mode collapse from training ProxT2I using reinforcement learning with Aesthetic score reward. Each row shows the images generated from a different prompt. Each column shows the images generated at different checkpoints throughout training, with the first representing the model before RL finetuning. As shown, although the RL algorithm initially improves the aesthetic quality, the outputs eventually collapse, converging to nearly identical images for different prompts.}
    \label{fig:aesthetic-collapse}
\end{figure*} 

\end{document}